\documentclass[sigconf]{acmart}

\usepackage{url}
\usepackage{bbm}
\usepackage{dsfont}
\usepackage{graphicx}
\usepackage{amsmath}
\usepackage{multirow}
\usepackage{longtable}
\usepackage{color}
\usepackage{rotating}
\usepackage{makecell}
\usepackage{array}
\usepackage{booktabs}
\usepackage{colortbl}
\usepackage{arydshln}

\usepackage{float}		
\usepackage{amsmath}        
\usepackage{graphicx}
\usepackage{multirow}
\usepackage{rotating}      
\usepackage{arydshln}		
\usepackage{wrapfig}       
\usepackage{array}         
\usepackage{enumitem}

\usepackage{tabu}  
\usepackage{bm}            
\usepackage{graphicx}
\usepackage{subfigure}
\usepackage{balance}

\usepackage{amsthm} 		

\definecolor{burntorange}{HTML}{CC5500}

\AtBeginDocument{%
  }

\setcopyright{acmlicensed}
\copyrightyear{2018}
\acmYear{2018}
\acmDOI{XXXXXXX.XXXXXXX}

\acmConference[Conference acronym 'XX]{Make sure to enter the correct
  conference title from your rights confirmation emai}{June 03--05,
  2018}{Woodstock, NY}
\acmISBN{978-1-4503-XXXX-X/18/06}

\begin{document}

\title{Powerformer: A Section-adaptive Transformer for Power Flow
Adjustment}

\author{Kaixuan Chen}
\authornote{Both authors contributed equally to this research.}
\author{Wei Luo}
\authornotemark[1]
\author{Shunyu Liu}
\authornote{Corresponding author: liushunyu@zju.edu.cn}
\affiliation{%
  \institution{State Key Laboratory of Blockchain \& Data Security, Zhejiang University}
  \city{Hangzhou}
  \country{China}
}

\author{Yaoquan Wei}
\author{Yihe Zhou}
\author{Yunpeng Qing}
\affiliation{%
  \institution{State Key Laboratory of Blockchain \& Data Security, Zhejiang University}
  \city{Hangzhou}
  \country{China}
}

\author{Quan Zhang}
\affiliation{%
  \institution{College of Electrical Engineering, Zhejiang University}
  \city{Hangzhou}
  \country{China}
}
\author{Yong Wang}
\affiliation{%
  \institution{State Grid Shandong Electric Power Company}
  \city{Shandong}
  \country{China}
}

\author{Jie Song}
\affiliation{%
  \institution{State Key Laboratory of Blockchain \& Data Security, Zhejiang University}
  \city{Hangzhou}
  \country{China}
}

\author{Mingli Song}
\affiliation{%
  \institution{State Key Laboratory of Blockchain \& Data Security, Zhejiang University}
  \city{Hangzhou}
  \country{China}
}

\renewcommand{\shortauthors}{Trovato et al.}

\begin{abstract}
  In this paper, we present a novel transformer architecture tailored for learning robust power system state representations, which strives to optimize power dispatch for the power flow adjustment across different transmission sections. 
  Specifically, our proposed approach, named \emph{Powerformer}, develops a dedicated section-adaptive attention mechanism, separating itself from the self-attention employed in conventional transformers.
  This mechanism effectively integrates power system states with transmission section information, which facilitates the development of robust state representations.
  Furthermore, by considering the graph topology of power system and the electrical attributes of bus nodes, we introduce two customized strategies to further enhance the expressiveness: graph neural network propagation and multi-factor attention mechanism.
  Extensive evaluations are conducted on three power system scenarios, including the IEEE 118-bus system, a realistic China 300-bus system, and a large-scale European system with 9241 buses, where \emph{Powerformer} demonstrates its superior performance over several popular baseline methods. The code is available at: https://github.com/Cra2yDavid/Powerformer
\end{abstract}



\received{20 February 2007}
\received[revised]{12 March 2009}
\received[accepted]{5 June 2009}

\maketitle

\section{Introduction}
\label{sec:introduction}

\begin{figure}
	\centering
	{\includegraphics[scale=0.28]{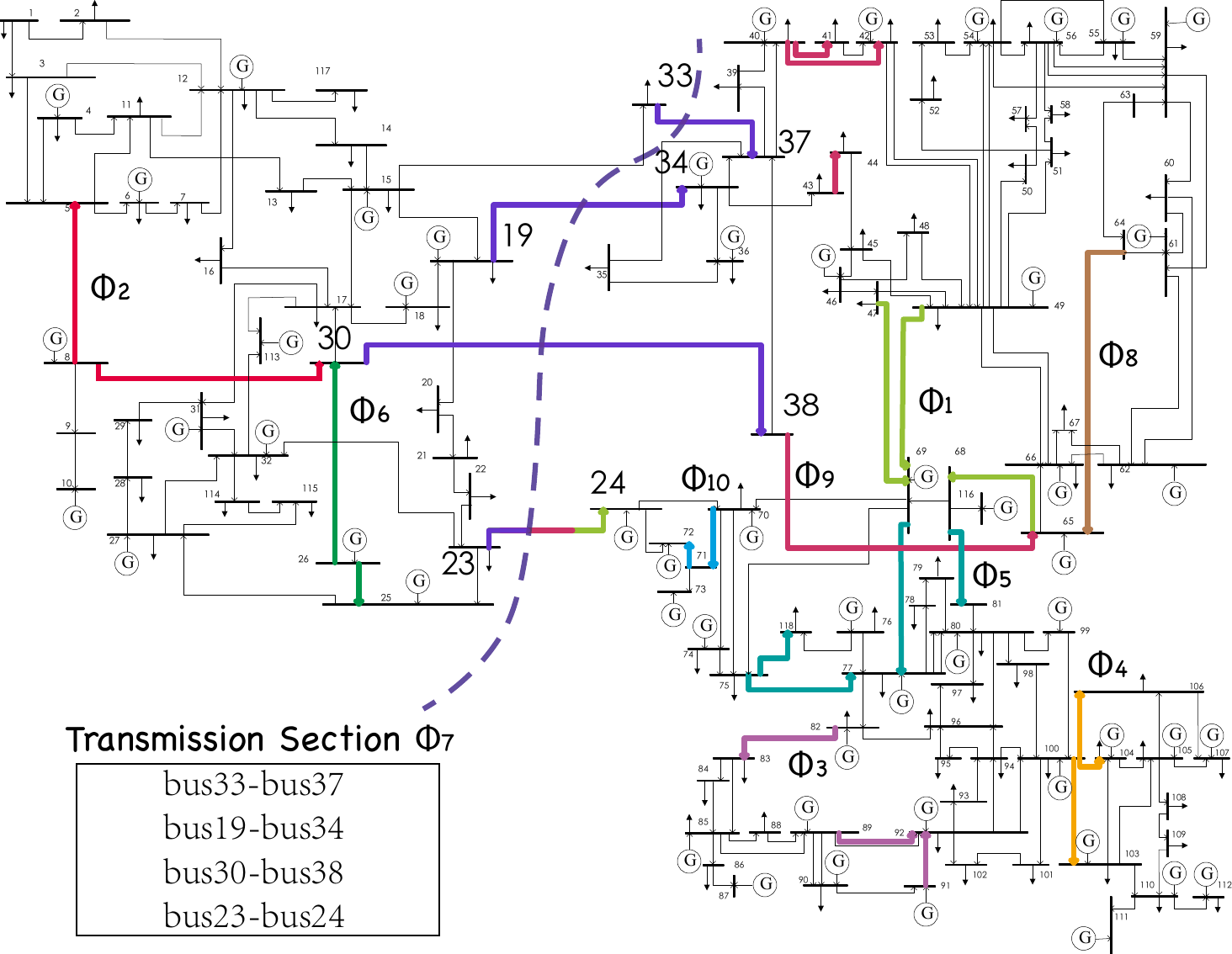}}
	\hspace{-0mm}
	\caption{Illustration of 10 transmission sections $\{\Phi_1,...,\Phi_{10}\}$ in the IEEE 118-bus system, where each section is represented by a set of transmission lines 
  that share the same color and are closely located electrically.
  An illustrative section $\Phi_7$ is the set of purple lines, i.e., $\Phi_7 = \,$\{\texttt{bus33-bus37}, \texttt{bus19-bus34}, \texttt{bus30-bus38}, \texttt{bus23-bus24}\}, that partitions the system into distinct left and right components. }
	\label{fig:case118_ti}
	\vspace{-0.3cm}  
\end{figure}

Power flow adjustment across the \emph{transmission sections} refers to manage and regulate the power flow between two system components or entities. 
As shown in Fig.~\ref{fig:case118_ti}, 
a transmission section consists of a group of transmission lines with the same color that have the same direction of active power flow and are closely located electrically.
\emph{In a real-world scenario, the power demand in China's northern area drastically increases due to the arrival of winter, and the power generation in this area falls insufficient, making it necessary to transfer power from the southern to the northern through the transmission section.} 
To ensure power system security margins, operators commonly employ the total transfer capability of the transmission section to monitor power flows and regulate the operational state~\cite{min2006total,liu2015iterative,jiang2022steady}.
Specifically, overloading a critical transmission section poses a serious threat to the system, 
potentially triggering a chain reaction of failures that may cause severe cascading blackouts.
Hence, the efficient management of power flow at transmission sections is paramount, serving as a fundamental pillar to ensure the stability, resilience, and reliability of the power system.

To address power flow adjustment across different transmission sections, power dispatch is an effective strategy that enables precise finetuning of generation outputs to optimize power flow distribution.
Traditional dispatch methods, which rely on the system model, can be primarily categorized into two distinct classes: optimization-based~\cite{capitanescu2011state} and sensitivity-based~\cite{dutta2008optimal} techniques.
Optimization-based methods utilize numerical techniques to transform the dispatch problem into a constrained programming problem, while sensitivity-based methods apply iterative calculations 
to find candidate generators by calculating sensitivity indices.
However, these model-based methods often encounter the computational burdens and struggle to effectively control unpredictable system coupling within a limited time. 
Thus, traditional methods are unsuitable for effectively managing real-time control in a complex and large-scale power system.

{
Recently, with the advancement of deep learning~\cite{lecun2015deep,shi2024scene,vaswani2017attention,shi2022foveated,shi2021foveated}, deep reinforcement learning (DRL) methods including DQN~\cite{MnihKSRVBGRFOPB15}, PPO~\cite{schulman2017proximal}, and A2C~\cite{Mnih2016AsynchronousMF} have gained attention for their potential in addressing various control challenges within power systems, such as voltage control~\cite{mu2023graph,yang2019two,cao2021deep}, economic dispatch~\cite{hu2023towards,dai2019distributed,li2021virtual}, and power flow adjustment~\cite{wu2023constrained,gao2021hybrid,wang2022stabilizing}.
} 
For the specific issue of power flow adjustment, 
these data-driven approaches effectively utilizes the robust capabilities of the pertained deep neural networks, enabling the direct learning of response actions in an end-to-end fashion, effectively tackling the high-dimensional control problems and providing a powerful solution for power flow adjustment.
However, among the existing DRL-based  methods~\cite{wu2023constrained,gao2021hybrid,wang2022stabilizing,wei2023agent,qinga2po},
there are two evident limitations in the state representations for transmission section power flow adjustment: 

\begin{enumerate}[leftmargin=*]
  \item \textbf{The absence of efficient awareness regarding transmission sections,} which hinders the DRL agent from making optimal action for distinct section adjustment.

  \item \textbf{The high coupled electrical factors and the ignored topology structure of power system,} ignificantly affect the extraction of the state characteristics.
\end{enumerate}

In this paper, we strive to develop an advanced transformer architecture for learning robust power system state representations, 
enabling the DRL agent to perform precise power dispatch strategies that stabilize the power system under various section settings. 
\emph{To tackle the first limitation,} 
we develop a section-adaptive attention mechanism to enable an efficient amalgamation of section information with the power system state, leveraging the comprehensive capabilities of the transformer architecture~\cite{vaswani2017attention,Zhu2023OnSE,zhang2022bootstrapping,ding2024freecustom} for the specific sectional system analysis. 
\emph{For the second limitation,} 
we begin by analyzing the graph topology of the power system, and employing graph neural networks \cite{jing2023efficient,Li2023MessagepassingST,yu2024multi,chen2023improving} 
to propagate features for embedding structural information into state representations.
Furthermore, 
we introduce a multi-factor attention mechanism to efficiently fuse each state factor, 
ensuring comprehensive and accurate extraction of system states.
Finally, we utilize the proposed \emph{Powerformer}, integrating the Dueling DQN~\cite{Wang2015DuelingNA} framework to achieve precise and efficient control of power flow across transmission sections, thereby leading to improvements in the overall system performance.

To the best of our knowledge, this is the first Transformer architecture that attempts to amalgamate the section information with state representation for transmission section power flow adjustment.  
Our main contributions can be summarized as follows: 
\begin{itemize}[leftmargin=*]
  \item We develop a section-adaptive attention mechanism that effectively integrates power system state features and section power flow information to achieve the knowledge amalgamation for different transmission sections.

  \item We introduce graph neural networks propagation alongside a multi-factor attention mechanism, specifically tailored to fuse the power system's graph topology information and effectively emerge electrical state factors.

  \item Extensive evaluation on three distinct power system cases, including the IEEE 118-bus system, a realistic China 300-bus system, and a large-scale 9241-bus European system, demonstrates our method's superiority with significantly improved performance.

\end{itemize}

\section{Preliminary}


\subsection{Problem Formulation}\label{problem_task}

Transmission section power flow adjustment involves the regulation and control of electrical power flow through the specific section or interface of a power transmission network. 
In a specific power system, there exist $N$ transmission sections $\{\Phi_1,\Phi_2,...,\Phi_N\}$, where each section $\Phi_i$ consists of several transmission lines that not only 
share the same direction of active power flow but also maintain close electrical proximity to one another~\cite{liu2023transmission}.
To accurately determine the total power flow of a transmission section $\Phi_i$, the combined power contributions are aggregated and meticulously calculated:

\begin{equation}
\begin{aligned}
    &\mathcal{P}_{\Phi_i} = \sum_{j \in \Phi_i} \mathcal{P}_{j} \quad \, {s.t.} \; \mathcal{P}_{\text{min}, \Phi_i} \le \mathcal{P}_{\Phi_i} \le \mathcal{P}_{\text{max}, \Phi_i}, \\
    &\mathcal{Q}_{\Phi_i} = \sum_{j \in \Phi_i} \mathcal{Q}_{j} \quad \,{s.t.} \; \mathcal{Q}_{\text{min}, \Phi_i} \le \mathcal{Q}_{\Phi_i} \le \mathcal{Q}_{\text{max}, \Phi_i},
\end{aligned}
\end{equation}
where $\mathcal{P}_{\Phi_i}$ and $\mathcal{Q}_{\Phi_i}$ represent the total active and reactive power flow in the $i$-th transmission section, and $\mathcal{P}_j$ and $\mathcal{Q}_j$ denote the active and reactive power flow in the $j$-th transmission line within the section $\Phi_i$. $\mathcal{P}_{\text{min}, \Phi_i}$, $\mathcal{P}_{\text{max}, \Phi_i}$, $\mathcal{Q}_{\text{min}, \Phi_i}$, $\mathcal{Q}_{\text{max}, \Phi_i}$ are the minimum and maximum allowable active and reactive power flow in the $i$-th transmission section. 
These limits are essential for ensuring the power system operates safely and efficiently, maintaining power flow within each transmission section at levels that prevent overloading and stability issues. 
To address this problem, power dispatch via DRL is a commonly employed strategy for regulating power system flow and maintaining safety margin constraints.


\subsection{Sectional Power System Network}
The power system topology, being non-Euclidean~\cite{chen2020covariance,Chen2022RiemannianLM}, is typically represented as a graph structure~\cite{chen2023hybrid,chen2023tele,chen2023improving}. Therefore, we organize power system data in the form of graph-structured data for learning representations.
In a power system graph network, nodes represent buses and edges symbolize transmission lines. 
Specifically, a power system network with section information can be defined as:

\vspace{0.1cm}
\noindent
\textbf{Definition 1} \emph{(Sectional Power System Network) A sectional power system network is defined as $\mathcal{G}=(\mathcal{V}, \mathcal{E}, \boldsymbol{H}, \boldsymbol{Z}_{\Phi})$, where $\mathcal{V}=\{v_1,...,v_n\}$ and  $\mathcal{E}=\{e_1,...,e_m\}$ represent the sets of $n$ nodes and $m$ edges, respectively. 
$\boldsymbol{H}=[\boldsymbol{h}_1,...,\boldsymbol{h}_n] \in\mathbb{R}^{d\times n}$ denotes the state feature matrix, where $\boldsymbol{h}_i$ denotes the $i$-th node. $\boldsymbol{Z}_{\Phi}$ is the representation of transmission section $\Phi$. Specifically, nodes' features are initialized with active power, reactive power, voltage magnitude, and phase angle. $\boldsymbol{Z}_{\Phi}$ is initialized using the active power on the $i$-th section.}

\vspace{0.1cm}
\noindent
To illustrate a sectional power system network, Fig.~\ref{fig:case118_ti} provides an example with 10 sections in the IEEE 118-case system. In this case, transmission section $\Phi_7$ is represented by the purple lines, which divides the system into left and right components or entities.

\begin{figure*}
	\centering
	{\includegraphics[scale=0.32]{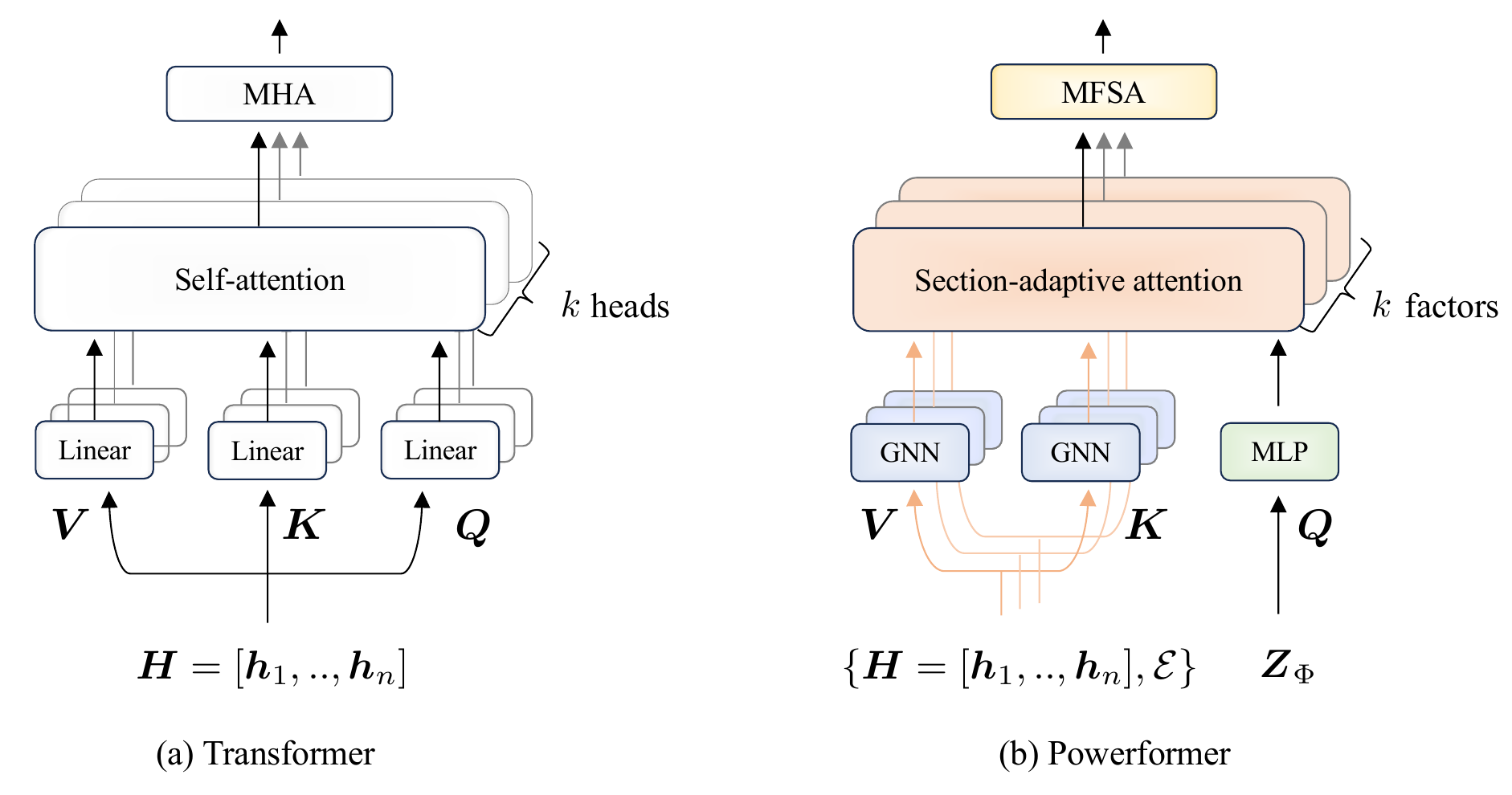}}
	\caption{The illustration of the Transformer and Powerformer architectures. (a) Transformer architecture with the multi-head self-attention mechanism (MHA). (b) Powerformer architecture with the multi-factor section-adaptive attention (MFSA). }
	\label{fig:powerformer}
\end{figure*}

\subsection{Transformer Architecture}
Numerous Pre-trained Language Models (PLMs) employ a multi-layer Transformer architecture \cite{vaswani2017attention} for the purpose of encoding text representations. 
In each layer of the Transformer, a multi-head self-attention mechanism is utilized to generate contextualized representations for individual text tokens.
Let $\boldsymbol{H}^{(l)}=[\boldsymbol{h}_1^{(l)},...,\boldsymbol{h}_n^{(l)} ]$ denote the output feature of the $l$-th Transformer layer, where $\boldsymbol{h}_i^{(l)} \in \mathbb{R}^{d\times 1}$ is the hidden representation of the text token at position $i$. Subsequently, in the $(l+1)$-th Transformer layer, the multi-head self-attention (MHA) is calculated as follows:

\begin{equation}
\text{MHA}(\boldsymbol{H}^{(l)}) = \underset{t=1}{\overset{k}{\Vert}} \text{head}^t(\boldsymbol{H}^{(l)}),
\end{equation}
\begin{equation}
\text{head}^t(\boldsymbol{H}^{(l)}) = \boldsymbol{V}_t^{(l)} \cdot \text{softmax}\left(\frac{{\boldsymbol{K}_t^{(l)\top} \boldsymbol{Q}_t^{(l)}}}{\sqrt{d}}\right),
\label{self_attention_representation}
\end{equation}
\begin{equation}
 \boldsymbol{Q}_t^{(l)} = \boldsymbol{W}_{Q,t}^{(l)} \boldsymbol{H}^{(l)},\; \boldsymbol{K}_t^{(l)} = \boldsymbol{W}_{K,t}^{(l)} \boldsymbol{H}^{(l)},\; \boldsymbol{V}_t^{(l)} = \boldsymbol{W}_{V,t}^{(l)} \boldsymbol{H}^{(l)},
\end{equation}
where $\boldsymbol{Q}_t^{(l)},\boldsymbol{K}_t^{(l)},\boldsymbol{V}_t^{(l)}$ are the representations of \texttt{Query}, \texttt{Key}, \texttt{Value} components in the $l$-th layer Transformer of the $t$-th attention head in MHA;
$\boldsymbol{W}_{Q,t},$ $\boldsymbol{W}_{K,t},$ $\boldsymbol{W}_{V,t}$ are the corresponding parameter matrices to be learned by the model. $k$ is the total number of attention heads and $\Vert$ is the concatenate operation. The symbol $\top$ represents the transpose of matrix, and $\cdot$ denotes the dot product operation. 
${{\boldsymbol{K}_t^{(l)\top} \boldsymbol{Q}_t^{(l)}}}/{\sqrt{d}}$ is the self-attention matrix of the $t$-th head at the $l$-th layer.
Fig.~\ref{fig:powerformer}a illustrates the traditional Transformer architecture.

\section{Method}

\subsection{Overall Architecture}
An overview of the Powerformer architecture is presented in Fig.~\ref{fig:powerformer}b, 
illustrating the crucial components for obtaining robust representations of sectional power system networks, such as State Factorization, Graph Neural Network (GNN) propagation, and Section-adaptive Attention, etc.
Different from the conventional Transformer, our architecture utilizes the set $\{\boldsymbol{H}, \boldsymbol{Z}_\Phi, \mathcal{E} \}$ as input, which consists of the state feature matrix, section power flow, and graph structure information.
The section-specific information $\boldsymbol{Z}_\Phi$ is utilized to learn \texttt{Query} representation, enabling the development of section-adaptive attention. 
To address the specific electrical factors on buses, we disentangle the system state representation into $k$ distinct factors to prevent feature coupling during propagation, and then incorporate the section-adaptive attention matrices to establish the multi-factor section-adaptive attention mechanism. 
Furthermore, to enhance the expressiveness of \texttt{Key}, \texttt{Value} propagation, we utilize the popular GNN architecture to embed the graph structure information $\mathcal{E}$ into state representations.

\subsection{Powerformer for Sectional Power System}

\noindent 
\textbf{State Factorization and GNN propagation.} Each node in the power system includes important state factors, such as active power, reactive power, voltage magnitude, and phase angle, providing comprehensive insights into the state characteristic of power system. 

If the state matrix is directly inputted into the propagation layer, it will result in the coupling of different factors, causing the redundancy in the learned features and ultimately diminishing the expressive power.
Thus, we begin by disentangling the input state feature into $k=4$ factors (i.e., active power, reactive power,
voltage magnitude, and phase angle),  
with the $l$-th layer's representation of the $t$-th factor denoted by $\boldsymbol{H}_t^{(l)}\in\mathbb{R}^{d\times n}$.

Furthermore, for the propagation of the \texttt{Key} and \texttt{Value} components in relation to the $t$-th factor, we utilize the widely-used GNN to incorporate the power system's graph topological structure for efficient updates:
\begin{equation}
  \boldsymbol{K}_t^{(l)} = \mathcal{F}_{K,t}^{(l)}(\boldsymbol{H}_t^{(l)},\mathcal{E}),\; \boldsymbol{V}_t^{(l)} = \mathcal{F}_{V,t}^{(l)} (\boldsymbol{H}_t^{(l)},\mathcal{E}),
  \label{key_value_propagation}
\end{equation}
where $\mathcal{F}_{K,t}^{(l)},\mathcal{F}_{V,t}^{(l)}$ represent the GNN propagation to update \texttt{Key} and \texttt{Value}: $\boldsymbol{K}_t^{(l)}, \boldsymbol{V}_t^{(l)}\in\mathbb{R}^{d\times n}$ of the $t$-th factor. 
Specifically, this propagation is implemented by Graph Isomorphism Network (GIN)~\cite{xu2018powerful}:
\begin{align}
	\boldsymbol{h}_{v,t}^{(l)} =   {\rm MLP}^{(l)}(( 1 + \epsilon^{(l)} ) \cdot \boldsymbol{h}_{v,t}^{(l-1)} +  \sum\nolimits_{u \in \mathcal{N}(v)} \boldsymbol{h}_{u,t}^{(l-1)}),
	\label{eq:gin}
\end{align}
where $\boldsymbol{h}_{v,t}^{(l)}$ represents the feature vector of node $v$ of the $t$-th factor at the $l$-th layer. $\mathcal{N}(v)$ refers to the neighborhood set of node $v$ based on the edge set $\mathcal{E}$. MLP is the Multi-layer Perceptron network.

\vspace{0.1cm}
\noindent
\textbf{Section-adaptive Attention (SA).} 
For section-adaptive representations, 
we first utilize the power flow information (i.e., active power) of the transmission lines on the current section to encode the section, and set the power flow information that is not within the section to zero. The resulting initialization representation is $\boldsymbol{Z}_\Phi$.
Then, we use a MLP network to learn the \texttt{Query} component:
\begin{equation}
  \boldsymbol{Q}^{(l)} = \mathcal{M}_{Q}^{(l)} (\boldsymbol{Z}_\Phi^{(l)}),
  \label{query_propagation}
\end{equation}
where $\mathcal{M}_{Q}^{(l)}$ is the employed MLP to update section representation. The updated \texttt{Query} is $\boldsymbol{Q}^{(l)}\in \mathbb{R}^{d\times 1}$.
Moreover, it is worth noting that the \texttt{Query} component in our architecture does not perform the state factorization.
Then, the section-adaptive attention without softmax function within the $t$-factor can be expressed as:
\begin{equation}
  \boldsymbol{a}_t^{(l)}=\text{SA}^t(\boldsymbol{K}_t^{(l)}, \boldsymbol{Q}^{(l)}) = \frac{{\boldsymbol{K}_t^{(l)\top} \boldsymbol{Q}^{(l)}}}{\sqrt{d}},
  \label{sa_representation}
\end{equation}
where the achieved $\boldsymbol{a}_t^{(l)}\in \mathbb{R}^{n\times 1}$ is the section-adaptive attention matrix of the $t$-th factor at the $l$-layer.


\vspace{0.1cm}
\noindent
\textbf{Multi-factor Section-adaptive Attention (MFSA).} 
In order to achieve an effective representation fusion of multiple factors, it is essential to effectively integrate the weights allocated to different factors across nodes.
Particularly, we recombine a new attention matrix, which is actually formed by combining different section-adaptive attention (Eq.\ref{sa_representation}) matrices, and then apply the softmax function on the dimensions of these factors:
\begin{equation}
  \widetilde{\boldsymbol{A}^{(l)}} = \text{softmax}(\boldsymbol{A}^{(l)}),
  \label{mfsa_soft_attention}
\end{equation}
\begin{equation}
  \boldsymbol{A}^{(l)}=[\boldsymbol{a}_1^{(l)},...,\boldsymbol{a}_t^{(l)},...,\boldsymbol{a}_k^{(l)}], \;\,
  t=1,...,k, 
  \label{mfsa_attention}
\end{equation}
where $\widetilde{\boldsymbol{A}^{(l)}} \in \mathbb{R}^{n\times k}$ is the resulting matrix by applying the softmax function along the factor's dimension. Finally, the Multi-factor Section-adaptive Attention (MFSA) mechanism:
\begin{equation}
  \text{MFSA}(\mathcal{G}) = \underset{t=1}{\overset{k}{\sum }}\, \boldsymbol{V}_t^{(l)} \odot  \widetilde{\boldsymbol{A}^{(l)}}_t
  \label{mfsa}
\end{equation}
where $\boldsymbol{V}_t^{(l)}$ and $\widetilde{\boldsymbol{A}^{(l)}}$ are obtained from Eq.~\ref{key_value_propagation} and \ref{mfsa_soft_attention}, $\widetilde{\boldsymbol{A}^{(l)}}_t\in \mathbb{R}^{n\times 1}$ the $t$-th column associated with $t$-th electrical factor. Symbol $\odot$ denotes the Hadamard product with broadcast mechanism. 
It is important to note that the left matrix side of the $\odot$ operation is in $\mathbb{R}^{d\times n}$, while the right matrix is in $\mathbb{R}^{n\times 1}$. The resulting matrix after the Hadamard product will also reside in the space $\mathbb{R}^{d\times n}$. At last, we use \textit{mean pooling} as a readout operation to get graph-level representation.

\subsection{The Implementation of Powerformer}
At the implementation stage, we merge the Eq.~\ref{sa_representation}, \ref{mfsa_soft_attention} and \ref{mfsa_attention} into Eq.~\ref{mfsa} in order to provide a more clear formulation of our proposed Powerformer architecture:
\begin{equation}
    \text{MFSA}(\mathcal{G}) = \underset{t=1}{\overset{k}{\sum}}\, \text{factor}^t(\boldsymbol{H}^{(l)}, \boldsymbol{Z}_\Phi^{(l)},\mathcal{E}),
\end{equation}
\begin{equation}
    \text{factor}^t(\boldsymbol{H}^{(l)}, \boldsymbol{Z}_\Phi^{(l)},\mathcal{E}) = \boldsymbol{V}_t^{(l)} \odot \text{softmax}\left(\boldsymbol{A}^{(l)}\right)_t,
    \label{factor_representation}
\end{equation}
\begin{equation}
  \boldsymbol{A}^{(l)}=[\boldsymbol{a}_1^{(l)},...,\boldsymbol{a}_t^{(l)},...,\boldsymbol{a}_k^{(l)}], \;\,
  \boldsymbol{a}_t^{(l)}=\frac{{\boldsymbol{K}_t^{(l)\top} \boldsymbol{Q}^{(l)}}}{\sqrt{d}},
  \label{mfsa_attention_final}
\end{equation}
\begin{equation}
  \begin{split}
    &\boldsymbol{Q}^{(l)} = \mathcal{M}_{Q}^{(l)} (\boldsymbol{Z}_\Phi^{(l)}),\\
    \boldsymbol{K}_t^{(l)} = \mathcal{F}_{K,t}^{(l)}&(\boldsymbol{H}_t^{(l)},\mathcal{E}),\;
    \boldsymbol{V}_t^{(l)} = \mathcal{F}_{V,t}^{(l)} (\boldsymbol{H}_t^{(l)},\mathcal{E}),  
  \end{split}
\end{equation}
where $\mathcal{M}_{Q}^{(l)}$, $\mathcal{F}_{K,t}^{(l)}$, $\mathcal{F}_{V,t}^{(l)}$ are the corresponding propagation function for the \texttt{Query}, \texttt{Key} and \texttt{Value} components. Notably,
our section-adaptive attention based representation (Eq.~\ref{factor_representation}) diverges from the self-attention based representation (Eq.~\ref{self_attention_representation}) by employing softmax on the attention matrix (Eq.~\ref{mfsa_attention_final}) that considers all factors, rather than the attention matrix of a single head. Furthermore, we employ the summation operation to combine diverse factor representations, rather than concatenation in MHA.

\begin{figure*}
  \centering
  \begin{minipage}{1.0\linewidth}
    \centering
    \subfigure[Concat]{\includegraphics[width=0.3\textwidth]{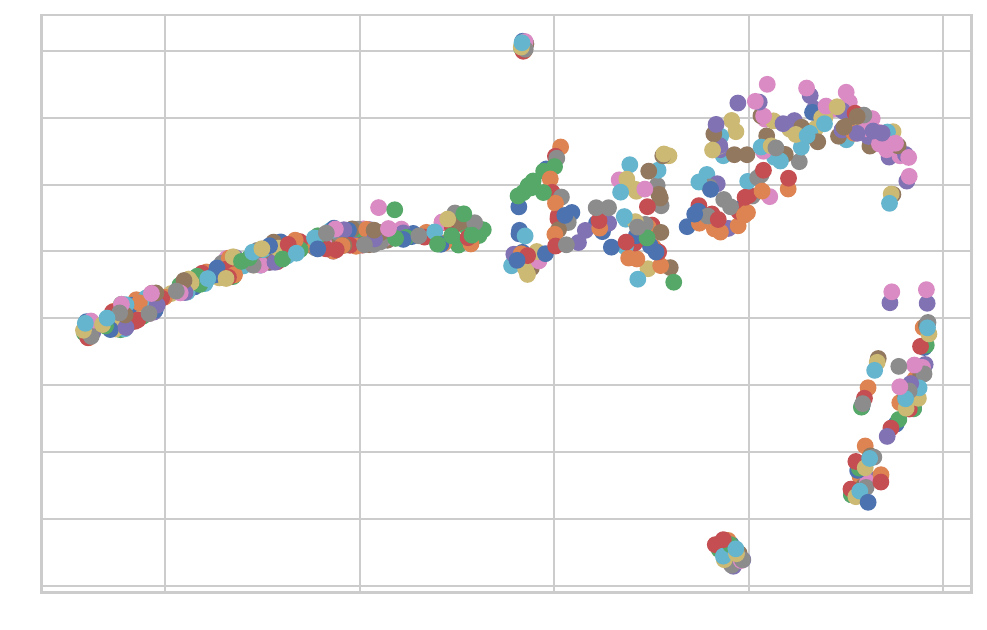}\label{fig:tsne_concat}}\hfil\hfil
    \subfigure[Attention]{\includegraphics[width=0.3\textwidth]{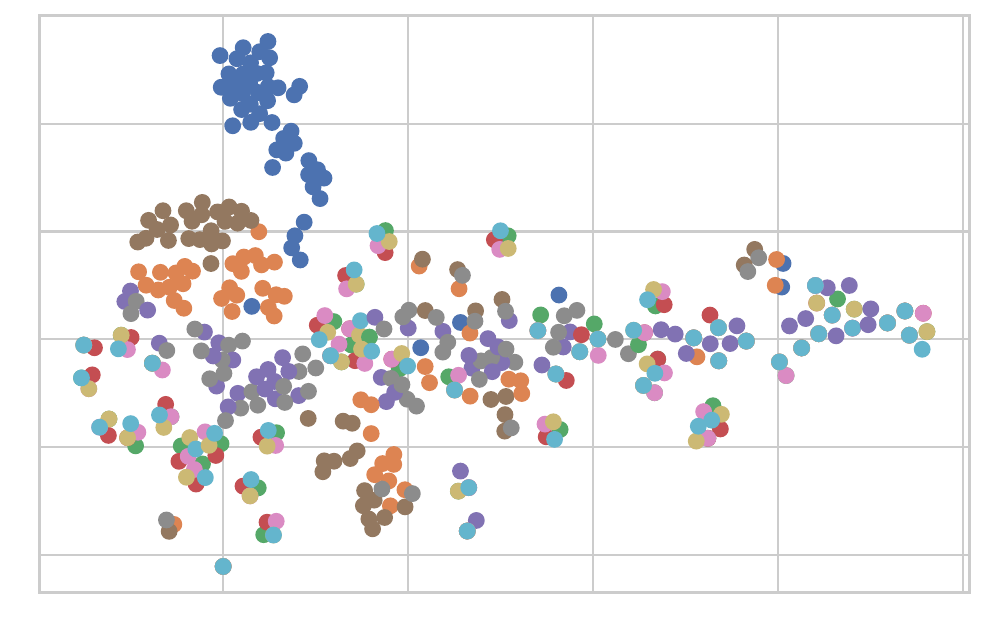}\label{fig:tsne_soft}}\hfil\hfil
    \subfigure[\textbf{Powerformer (Ours)}]{\includegraphics[width=0.3\textwidth]{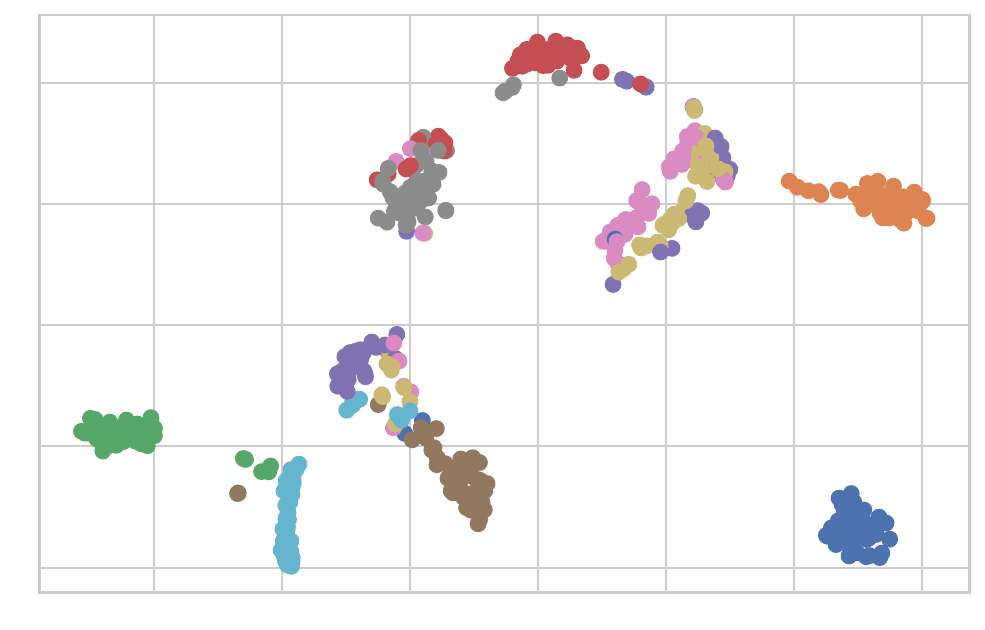}\label{fig:tsne_factor}}\\
    \caption{The visualization highlights the distribution of Concat, Attention, and Powerformer operations for combining state and 10 types of transmission section representations on the IEEE 118-bus system.}
    \vspace{0.2cm}
    \label{fig:vis1}
  \end{minipage}

  \begin{minipage}{1.0\linewidth}
    \centering
    \normalsize
    \renewcommand{\arraystretch}{1.0} 
    \captionof{table}{{The performance of our method and baselines on three different power systems is evaluated based on an average of 5 trials. The performance differences between our method and baselines are indicated in parentheses, with better performance denoted by a higher test success rate and lower test economic cost.}}
    \label{tab:representation_comparison}
    \resizebox{1.0\textwidth}{!}{%
    \begin{tabular}{@{}lcccccc@{}}
      \toprule
      \multicolumn{1}{l}{\multirow{2}{*}{{Method}}} 
      & \multicolumn{2}{c}{{118-bus System}} 
      & \multicolumn{2}{c}{{300-bus System}} 
      & \multicolumn{2}{c}{{9241-bus System}} 
      \\ \cmidrule(l){2-3} \cmidrule(l){4-5} \cmidrule(l){6-7} 
      & \multicolumn{1}{c}{{Success Rate (\%)}} & \multicolumn{1}{c}{{Economic Cost (\$)}} 
      & \multicolumn{1}{c}{{Success Rate (\%)}} & \multicolumn{1}{c}{{Economic Cost (\$)}}  
      & \multicolumn{1}{c}{{Success Rate (\%)}} & \multicolumn{1}{c}{{Economic Cost (\$)}} \\ \midrule
      Concat   
      & 68.44{\normalsize{~(-28.17)}}     & 626,215{\;\normalsize{~(+4,017)}} & 57.66{\normalsize{~(-37.14)}}
      & 992,048{\normalsize{~(+22,366)}}  & 37.57{\;\normalsize{~(-5.29)}}    & 324,485{\normalsize{\;\;\;\;\;\;\;~(-2)}}            \\ \specialrule{0em}{1pt}{1pt}
      Attention        
      & 85.23{\normalsize{~(-11.38)}}     & 625,128{\;\normalsize{~(+2,930)}} & 85.30{\;\normalsize{~(-9.50)}} 
      & 972,573{\;\normalsize{~(+2,891)}} & 31.13{\normalsize{~(-11.73)}}     & 325,152{\normalsize{\;\;\;~(+665)}}            \\ \specialrule{0em}{1pt}{1pt}
      \textbf{Powerformer (Ours)}              
      & \textbf{98.19}              & {\textbf{622,198}}            & \textbf{97.36} 
      & {\textbf{969,682}}            & \textbf{68.25}              & {\textbf{324,487}}          \\ \bottomrule
      \end{tabular}%
  }
\end{minipage}
\end{figure*}

\noindent
\textbf{Joint Training.} 
To deliver the final dispatch policy based on the proposed Powerformer framework, we adopt the Dueling Deep Q Network~(DQN) method\footnote{The reason for selecting Dueling DQN is provide in Appendix~\ref{dueling_dqn_selecting}. Action space and reward function are placed in Appendix~\ref{action_space}.} to train a deep reinforcement learning agent. Specifically, we introduce a Q-network $Q_\theta$  parameterized by $\theta$, which takes the state representation $\text{MFSA}(\mathcal{G})$ as input and estimates the Q-values for each possible dispatch action $d$. The dispatch policy with respect to $Q_\theta$ is defined as $\pi_Q(\mathcal{G})=\arg\max_{d\in\mathcal{D}}Q_\theta(\text{MFSA}(\mathcal{G}),d)$, where $\mathcal{D}$ is the dispatch action space.

During training, the agent first interacts with the environment to generate episodes using the $\epsilon$-greedy policy based on the the current Q-network $Q_\theta$. The transition tuples $\{\mathcal{G}, d, r, \mathcal{G}'\}$ encountered during training are stored in the replay buffer $\mathcal{B}$~\cite{MnihKSRVBGRFOPB15}, where $r$ is the environmental reward. 
Then the Q-network is trained using mini-batch gradient descent on the temporal difference loss as:
\begin{align}
\mathcal{L} = \mathbb{E}_{\mathcal{B}}\Big[Q_\theta(\text{MFSA}(\mathcal{G}), d) - (r + \gamma \cdot \max_{d'} Q_{\hat{\theta}}(\text{MFSA}(\mathcal{G}'), d'))\Big]^2,
\end{align}
where the tuples $\{\mathcal{G}, d, r, \mathcal{G}'\}$ are sampled from the replay buffer $\mathcal{B}$. DQN uses the target network to make the optimization procedure more stable. The target value $Q_{\hat{\theta}}$ depend on previous network parameters but this dependency is ignored during backpropagation.

\subsection{Discussions}
\noindent

Multi-channel graph transformers broadly apply self-attention mechanisms to enhance model performance, our architecture introduces a novel section-adaptive attention mechanism that is adaptive to the fluctuating demands in power system operations. 
In this subsection, we presents a brief discussion and analysis focusing on three aspects: section-adaptive attention mechanism, state factorization, and GNN integration with Transformers. \\
\noindent
\textbf{Section-adaptive Attention Mechanism.} Traditional attention-mask techniques primarily focus on preventing the model from accessing future information during training. However, these techniques typically rely on self-attention, which limit their ability to integrate heterogeneous transmission section information. Our section-adaptive approach differs in that it dynamically optimizes attention distribution based on the current section power flow. This allows for a more sophisticated adaptation to the varying demands of the system, enhancing decision-making for resource allocation.\\
\noindent
\textbf{State Factorization.} The state factorization aims to decouple the input into the clear electrical factors (i.e., active power, reactive power, voltage magnitude, and phase angle). They each possess explicit physical meaningful information. This is crucial for understanding the contribution of different factors to the state representation of the power system, which also enhances the interpretability of the achieved representation. As defined in Eq. \ref{mfsa} and Eq. \ref{factor_representation}, our softmax operates between different factors rather than different nodes.
In MHA, the electrical factors are coupling together, making it challenging for each individual attention head to interpret distinct electrical factor significances.\\
\noindent
\textbf{GNN integration with Transformers.} (1) For the task of transmission section power system adjustment, it is often to adjust the generators that are close to the section. Therefore, local information is important for the current section adjustment, and we need to use GNN to capture local information. However, as the power flow is adjusted, the overall operating conditions of the power system changes, and it also affects electrical equipment that is far away. Thus, we need to use Transformer to capture global information. 
(2) By leveraging the capability of GNN, our approach dynamically adapts to topological changes such as maintenance, upgrades, or unforeseen emergencies. For example, when a three-phase short circuit occurs on a transmission line, GNN can accurately capture the changes in local information. This helps to maintain robustness against real-world challenges.

For more detailed discussion, please refer to the Appendix \ref{detailed_analysis}.

\begin{figure*}
  \centering
  \begin{minipage}{1.0\linewidth}
    \centering
    \subfigure{\includegraphics[scale=0.26]{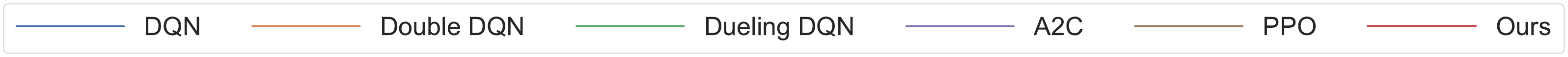}}
    \vspace{-0.1cm}
    \\
    \addtocounter{subfigure}{-1}

    \subfigure[IEEE 118-bus System (10-section Task)]{\includegraphics[width=0.3\textwidth]{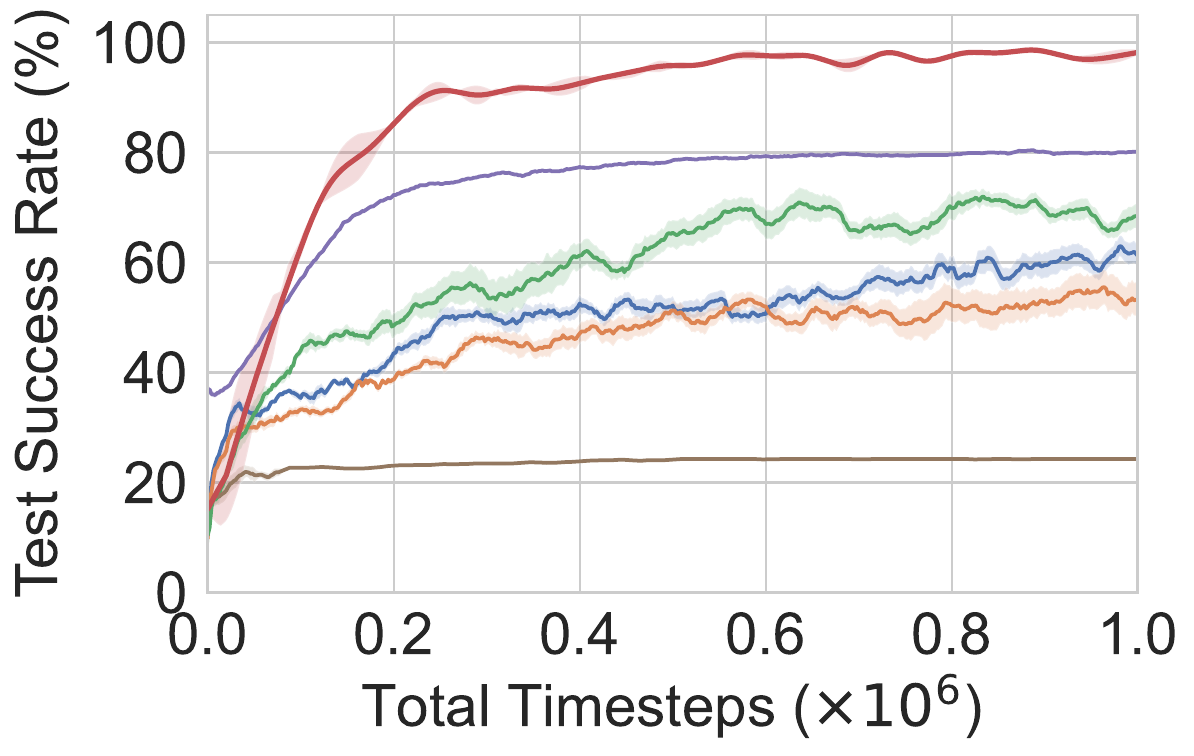}\label{fig:task10-118}} \hfil
    \subfigure[Realistic 300-bus System (10-section Task)]{\includegraphics[width=0.3\textwidth]{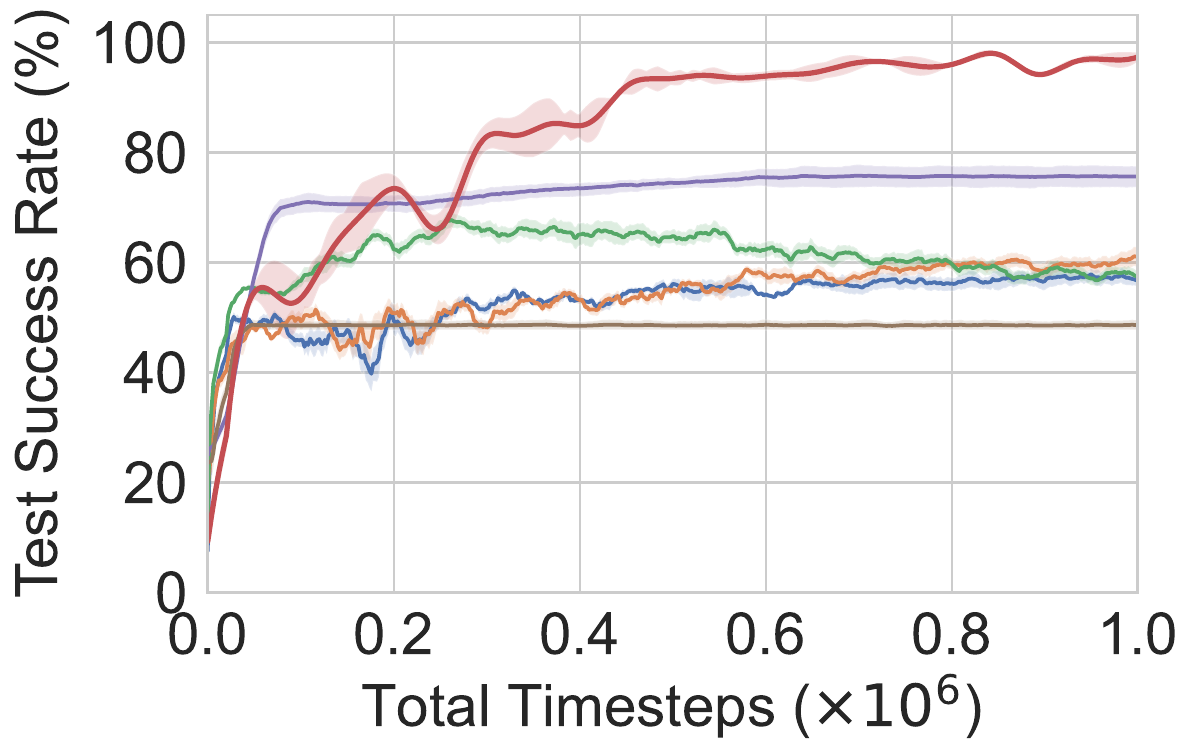}\label{fig:task10-300}} \hfil
    \subfigure[European 9241-bus System (10-section Task)]{\includegraphics[width=0.29\textwidth]{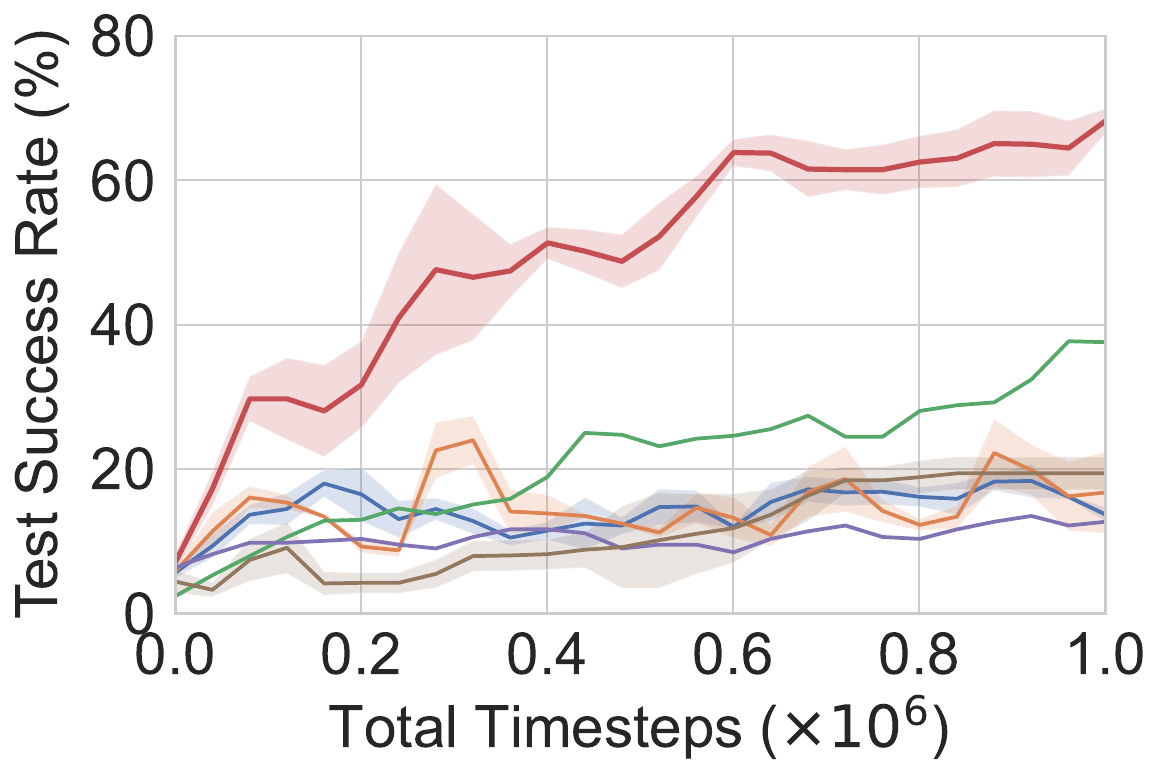}\label{fig:task10-9241}}
    \caption{The learning curves of all methods for 10-section task on three power systems. The experimental results use the median performance and one standard deviation (shaded region) over 5 random seeds to ensure a fair comparison.}
    \label{fig:learning_curve}
  \end{minipage}
  \\\vspace{0.2cm}
  \begin{minipage}{1.0\linewidth}
    \centering
    \normalsize    
  \renewcommand{\arraystretch}{1.0} 
    \captionof{table}{{The performance of our method and baselines for 10-section tasks on three different power systems is evaluated based on an average of 5 trials. The performance differences between our method and baselines are indicated in parentheses, with better performance denoted by a higher test success rate and lower test economic cost.}}
    \label{tab:total}
    \resizebox{0.98\textwidth}{!}{%
    \begin{tabular}{@{}lcccccc@{}}
      \toprule
      \multicolumn{1}{l}{\multirow{2}{*}{{Method}}} 
      & \multicolumn{2}{c}{{118-bus System~(10-section Task)}} 
      & \multicolumn{2}{c}{{300-bus System~(10-section Task)}} 
      & \multicolumn{2}{c}{{9241-bus System~(10-section Task)}} 
      \\ \cmidrule(l){2-3} \cmidrule(l){4-5} \cmidrule(l){6-7} 
      & \multicolumn{1}{c}{{Success Rate (\%)}} & \multicolumn{1}{c}{{Economic Cost (\$)}} 
      & \multicolumn{1}{c}{{Success Rate (\%)}} & \multicolumn{1}{c}{{Economic Cost (\$)}}  
      & \multicolumn{1}{c}{{Success Rate (\%)}} & \multicolumn{1}{c}{{Economic Cost (\$)}} \\ \midrule
      DQN          
      & 61.23{\normalsize{~(-36.96)}}     & 626,808{\;\normalsize{~(+4,610)}} & 56.74{\normalsize{~(-40.62)}} 
      & 991,762{\normalsize{~(+22,080)}}  & 13.67{\normalsize{~(-54.58)}}     & 324,658{\normalsize{\;\;\;~(+164)}}           \\  \specialrule{0em}{1pt}{1pt}
      Double DQN        
      & 53.28{\normalsize{~(-44.91)}}     & 638,079{\normalsize{~(+15,881)}}  & 61.24{\normalsize{~(-36.12)}} 
      & 990,126{\normalsize{~(+20,444)}}  & 16.75{\normalsize{~(-42.14)}}     & 324,792{\normalsize{\;\;\;~(+305)}}           \\ \specialrule{0em}{1pt}{1pt}
      Dueling DQN      
      & 68.44{\normalsize{~(-29.75)}}     & 626,215{\;\normalsize{~(+4,017)}} & 57.66{\normalsize{~(-39.70)}}
      & 992,048{\normalsize{~(+22,366)}}  & 37.57{\normalsize{~(-30.68)}}    & 324,485{\normalsize{\;\;\;\;\;\;\;~(-2)}}            \\ \specialrule{0em}{1pt}{1pt}
      A2C                
      & 80.14{\normalsize{~(-18.05)}}     & 626,703{\;\normalsize{~(+4,505)}} & 75.61{\normalsize{~(-21.75)}}
      & 976,467{\;\normalsize{~(+6,785)}} & 12.70{\normalsize{~(-55.55)}}     & 325,000{\normalsize{\;\;\;~(+513)}}           \\ \specialrule{0em}{1pt}{1pt}
      PPO               
      & 24.28{\normalsize{~(-73.19)}}     & 596,315{\normalsize{~(-25,883)}}  & 48.61{\normalsize{~(-48.75)}} 
      & 930,581{\normalsize{~(-39,101)}}  & 19.40{\normalsize{~(-48.85)}}     & 325,549{\normalsize{~(+1,062)}}          \\ \specialrule{0em}{1pt}{1pt}
      \cdashline{1-7}[1pt/1pt]
      {OPF}                        
      & 72.25{\normalsize{~(-25.94)}}     & 606,259{\normalsize{~(-15,939)}}  & 50.56{\normalsize{~(-46.80)}} 
      & 953,349{\normalsize{~(-16,333)}}  & 51.32{\normalsize{~(-16.93)}}    & {321,586{\normalsize{~(-2,901)}}}              \\ \midrule
      \textbf{Powerformer~(Ours)}              
      & \textbf{98.19}              & {\textbf{622,198}}            & \textbf{97.36} 
      & {\textbf{969,682}}            & \textbf{68.25}              & {\textbf{324,487}}          \\ \bottomrule
      \end{tabular}%
  }
\end{minipage}
\end{figure*}

\begin{figure*}
  \centering
  \begin{minipage}{1.0\linewidth}
    \centering
    \includegraphics[width=0.9\textwidth]{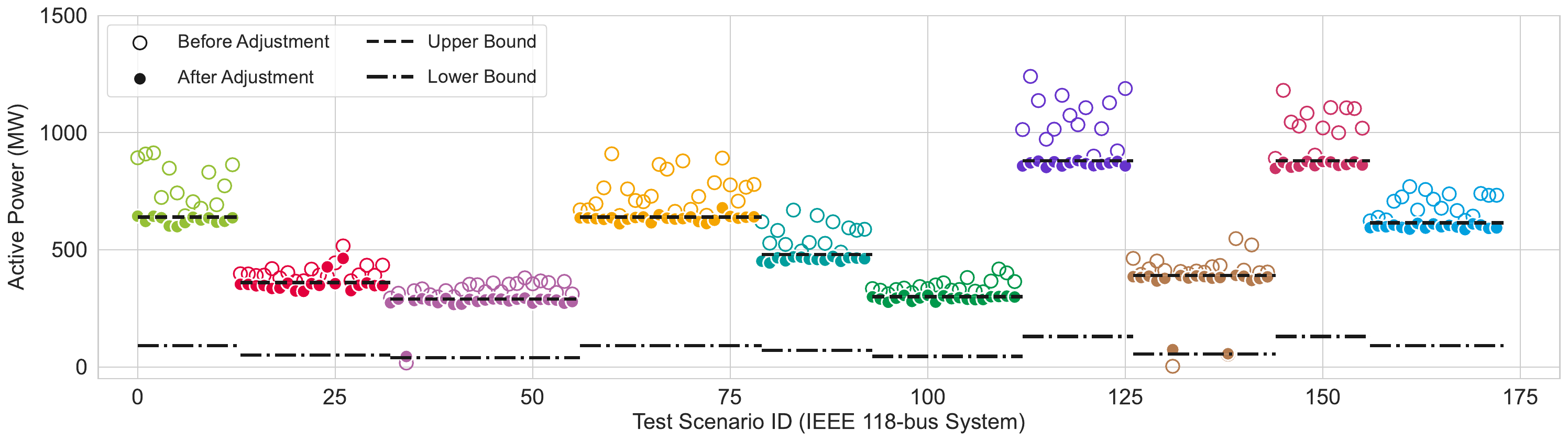}\label{fig:118pf}
    \caption{The visualization effectively emphasizes the active power of all scenario samples, both before and after adjustment, across all ten transmission sections of the IEEE 118-bus system using our method. Among them, 170 scenarios have been successfully adjusted into the capacity of the transmission section, i.e., 98.19\% test success rate reported in Table~\ref{tab:representation_comparison} and \ref{tab:total}.}
    \vspace{0.3cm}
    \label{fig:vis2}
  \end{minipage}
  \begin{minipage}{1.0\linewidth}
    \centering
    \includegraphics[width=0.9\textwidth]{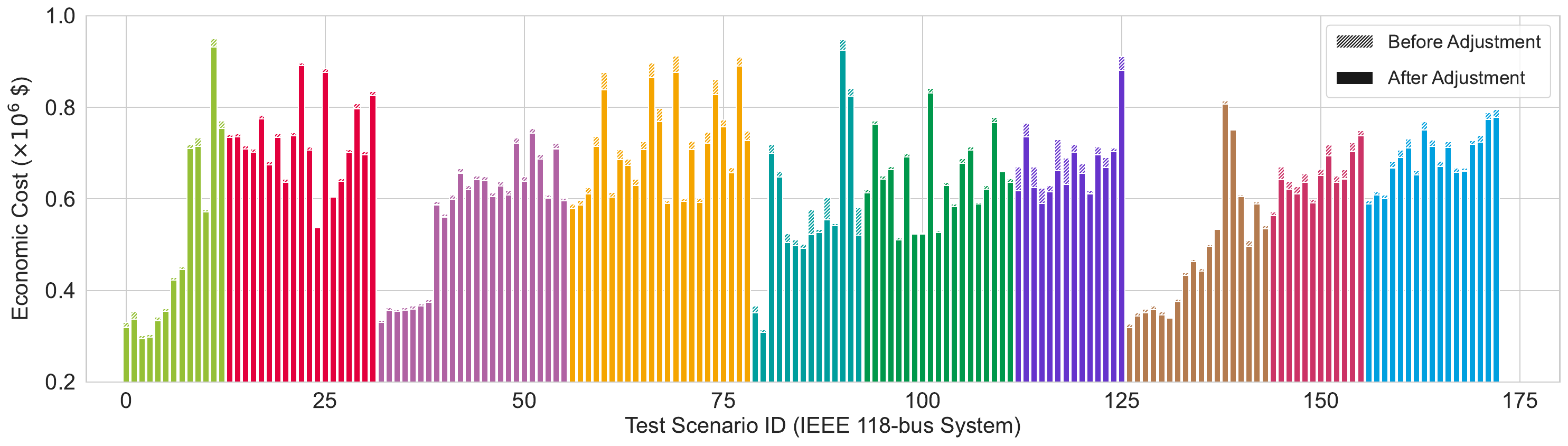}\label{fig:118ce}
    \caption{The visualization effectively emphasizes the economic cost of all scenario samples, both before and after adjustment, across all ten transmission sections of the IEEE 118-bus system using our method. 
    The economic costs of almost all samples have decreased, without any instances of cost increase.}
    \label{fig:vis3}
  \end{minipage}
\end{figure*}

\section{Experiments}

In this section, we first provide the details of experimental setting. 
Subsequently, we conduct a comparative analysis of our method with commonly used state representation learning methods, as well as several popular DRL-based adjustment baselines~\cite{MnihKSRVBGRFOPB15,Hasselt2015DeepRL,Wang2015DuelingNA,Mnih2016AsynchronousMF,schulman2017proximal,dommel1968optimal} and traditional Optimal Power Flow (OPF)~\cite{dommel1968optimal}.
Meanwhile, we use a visualized approach to display the section power flow adjustment, economic cost, and required inference time by using Powerformer adjustments.
Finally, we conduct ablation studies to enhance a comprehensive understanding of our proposed method.

\subsection{Experimental Setting}
\noindent
\textbf{Scenario Generation.}
To obtain adequate scenarios for training and testing purposes, we first employ two small-scale systems, namely the IEEE 118-bus system and a realistic China 300-bus system.
Then, we adopt a very large European 9241-bus system from the PEGASE project into our operations~\cite{Josz2016ACPF}.
The detailed scenario generation is provided in Appendix \ref{Scenario_Generation_supp}.

\vspace{0.1cm}
\noindent
\textbf{Compared Methods.}
We first consider two basic architectures of state representations for sectional power system:
\emph{1) Concat architecture.} We employ a MLP network that takes the concatenated vector of the section encoding vector and state feature vector as input.
\emph{2) Attention architecture.} We utilize the soft attention network, which allows agent to weight different network modules based on the section representation. Furthermore, the proposed Powerformer is compared with several popular DRL methods, including \emph{Deep Q Network}~(DQN)~\cite{MnihKSRVBGRFOPB15}, \emph{Double DQN}~\cite{Hasselt2015DeepRL}, \emph{Dueling DQN}~\cite{Wang2015DuelingNA}, \emph{Advantage Actor Critic}~(A2C)~\cite{Mnih2016AsynchronousMF}, \emph{Proximal Policy Optimization}~(PPO)~\cite{schulman2017proximal} and traditional \emph{Optimal Power Flow} (OPF)~\cite{dommel1968optimal}. Hyperparameter settings are provided in Appendix \ref{parameter_setting_supp}.

\vspace{0.1cm}
\noindent
\textbf{Evaluations.}
The evaluation of comparison analysis incorporates metrics such as the test success rate, test economic cost, and inference time.  
{
The economic cost is defined as:
\begin{equation}
  \begin{aligned}
      &\mathcal{R}(\text{MFSA}(\mathcal{G}),d) = \sum_{i =1}^{N_G} (\alpha_i (\mathcal{P}_{i}^G)^2+\beta_i \mathcal{P}_{i}^G +\lambda_i)
  \end{aligned}
\end{equation}
where $\alpha_i, \beta_i$ and $\lambda_i$ are the generation cost coefficients of the generator $i$. $\mathcal{P}_{i}^G$ is the active power production of the generator $i$. $N_G$ is the number of the generators.
}

\vspace{0.1cm}
\noindent
\textbf{Task Setting.} 
In the context of power flow adjustment across various transmission sections, multi-task learning~\cite{zhang2018overview,Zhou2023AutomaticTR} is suitable to manage power dispatch for simultaneously addresses different scenarios within the power system, where each section's power flow adjustment treated as a distinct task. 
In the main manuscript, we establish the 10-section power flow adjustment task on three distinct power systems. More task scenarios are considered in Appendix~\ref{total_section_task_supp}.

\subsection{Comparison with State Representations}

In this subsection, we aim to compare our novel Powerformer with two commonly used representation learning methods (concat and attention mechanism).
By conducting a comprehensive analysis using the same reinforcement learning framework of Dueling DQN~\cite{Wang2015DuelingNA}, we aim to clearly showcase the remarkable effectiveness and superiority of our proposed Powerformer method.

The performance results presented in Table~\ref{tab:representation_comparison} showcases a comparison between the Powerformer and two fundamental state encoding methods—Concat and Attention—across three different power system cases. The adopt evaluation is based on two metrics: success rate and economic cost, averaged over five trials.
Among these results, it is worth noting that the proposed Powerformer demonstrates the exceptional success rates of 98.19\%, 97.36\%, and 68.25\% respectively. This is particularly evident in its remarkable performance when applied to large-scale power systems such as the European 9241-bus system. 
In terms of economic cost, Powerformer tends to be more efficient or on par with the other methods. This suggests that Powerformer can achieve higher performance without incurring a significant test economic cost, which is a crucial factor to consider in practical applications.

Furthermore, we utilize the t-SNE toolkit to visually illustrate the distribution of state representations for 10 distinct sections on IEEE 118-bus system in Fig.~\ref{fig:vis1}.
Our proposed Powerformer approach can effectively cluster 10 different section representations into 10 central points, while the other two methods cannot accomplish.
This also explains the poor performance of the other two state encoding methods, as they couple the transmission section representation and different electrical factors of node features together~(Fig. \ref{fig:tsne_concat} and \ref{fig:tsne_soft}), making the agent impossible to execute specific action.

\subsection[short]{Comparison with Adjustment Baselines}

To demonstrate the superiority of our method in efficiently and effectively adjusting transmission section power flow, we thoroughly analyze and compare our method with various popular approaches in deep reinforcement learning (DRL) framework, as well as a well-established traditional OPF adjustment method. 

As shown in Fig.~\ref{fig:learning_curve}, 
our proposed method has successfully achieved the best test success rate among all considered DRL architectures on all three power systems. 
We can analyze and compare the superiority of our method based on the difficulty levels of task, especially in the more challenging large-scale power system case (i.e., Fig.~\ref{fig:task10-9241}), where the effectiveness of our method is highlighted.
This is because our method is able to learn better representations of sectional power system states, thus avoiding the coupling of state features. 
By comparative observing the significant differences on three power systems, we can conclude that when the system case is simple, the task is also relatively simple. As a result, the superiority of our method is not readily apparent.

In Table~\ref{tab:total}, we provide the comparative results of success rates and economic costs pertaining to power flow adjustment tasks.
For the IEEE 118-bus system, 
Powerformer outperforms all other methods with a success rate of 98.19\%, significantly surpassing A2C (80.14\%) with a substantial margin. 
For the China 300-bus system,
Powerformer continues to exhibit a satisfying level of performance in comparison to the baseline approaches, accomplishing an impressively high test success rate of 97.36\%.  
For the European-9241 system, 
our method consistently demonstrates exceptional performance, achieving an impressive success rate of 68.25\%; on the other hand, alternative DRL methods fail to surpass the 40\% threshold. 
Additionally, our method consistently demonstrates its exceptional ability to achieve relatively low economic costs across all scenarios.

\subsection{Visualization Analysis}

To better understand the superiority of the proposed Powerformer, we present illustrations that depict active power and economic cost across the 10 transmission sections of IEEE 118-bus system.

\noindent
\textbf{Active Power Comparison.} By comparing the active power of all scenario samples before and after Powerformer adjustment across all ten transmission sections on the IEEE 118-bus system, we can find that 170 test scenarios are successfully adjusted into the pre-determined range. Among these cases shown in Fig.~\ref{fig:vis2}, the majority involve adjusting the situation where the power flow exceeds the limit to align with the range of the upper bound, thereby guaranteeing the safe operation of the power system and ensuring its long-term stability.

\noindent
\textbf{Economic Cost Comparison.}
By comparing the economic cost of all scenario samples before and after Powerformer adjustment on all ten transmission sections of the IEEE 118-bus system, we can find that the economic cost of all test samples has been reduced. As shown in Fig.~\ref{fig:vis3}, after performing further adjustment through Powerformer, the economic cost of nearly every sample has been reduced, with no instances of cost increase. Hence, it can be inferred that our method reliably aims to reduce economic cost.

\begin{figure}
  \centering
  \begin{minipage}{1.0\linewidth}
      \centering
      \subfigure{\includegraphics[width=1.0\textwidth]{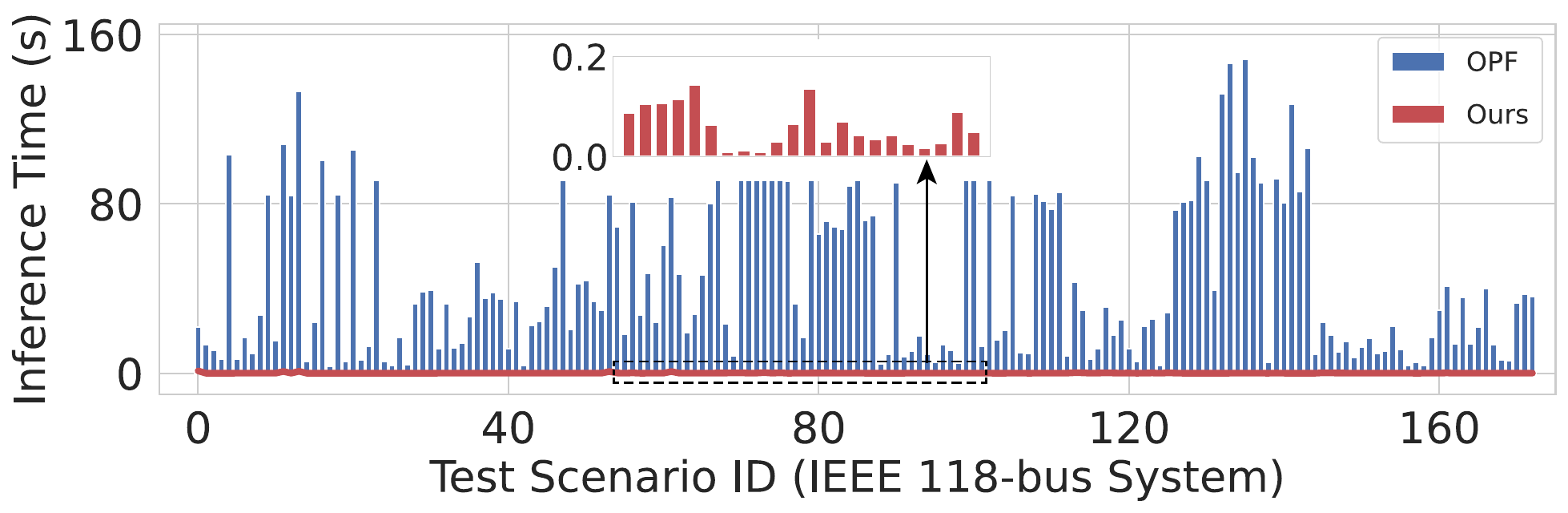}}
      \caption{The inference time comparison of total 173 scenarios on the IEEE 118-bus system. Our method is significantly faster than OPF in all 173 scenarios. }
      \label{fig:inferencetime}
  \end{minipage}
  \begin{minipage}{1.0\linewidth}
      \centering
      \vspace{0.3cm}
      \normalsize
      \renewcommand{\arraystretch}{1.0} 
      \captionof{table}{The inference time of our method and OPF method on three different power systems. Our method's inference time is significantly lower than OPF on three datasets.}
      \label{tab:inferencetime}
      \resizebox{1.0\textwidth}{!}{%
      \begin{tabular}{@{}lccc@{}}
      \toprule
      & \multicolumn{1}{c}{\textbf{118-bus System}} & \multicolumn{1}{c}{\textbf{300-bus System}} & \multicolumn{1}{c}{\textbf{9241-bus System}}\\ \midrule
      OPF                        & 44.822 $\pm$ 38.459      & 47.668 $\pm$  28.230    & 1912.608 $\pm$  1636.746        \\ \specialrule{0em}{1pt}{1pt}
      Powerformer~(Ours)              & 0.078  $\pm$ 0.151    & 0.121   $\pm$   0.176  & 0.585   $\pm$   0.602     \\ \bottomrule
      \end{tabular}%
      }
  \end{minipage}
\end{figure}

\subsection{Inference Time Analysis}
    

In real-time applications, inference time plays a critical role in large-scale systems, as exemplified by the European 9241-bus system.
In such scenarios, power flow adjustment becomes more complex due to the numerous variations that need careful consideration and strategic decision-making.
To better understand the superiority of our method in terms of inference speed, we provide Fig.~\ref{fig:inferencetime} and Table~\ref{tab:inferencetime} to demonstrate the advantages of our approach when compared to traditional OPF method used in practical scenarios.
As shown in Fig.~\ref{fig:inferencetime}, 
the experimental results demonstrate that, when compared to the traditional OPF method, the time required to successfully achieve the predefined range for each individual test is significantly reduced across a comprehensive set of test scenarios in the IEEE 118-bus system.
Moreover, it is notable that our method exhibits a noteworthy reduction in averaged inference time across three different power systems, as evidenced by the compelling numerical statistics presented in Table~\ref{tab:inferencetime}.
As a result, 
our proposed approach exhibits significantly improved time efficiency compared to the traditional OPF method, particularly when dealing with larger system sizes, rendering the OPF method impractical.


\subsection{Ablation Study}
\begin{figure}
  \centering
  \begin{minipage}{1.0\linewidth}
    \centering
    {\includegraphics[width=0.75\textwidth]{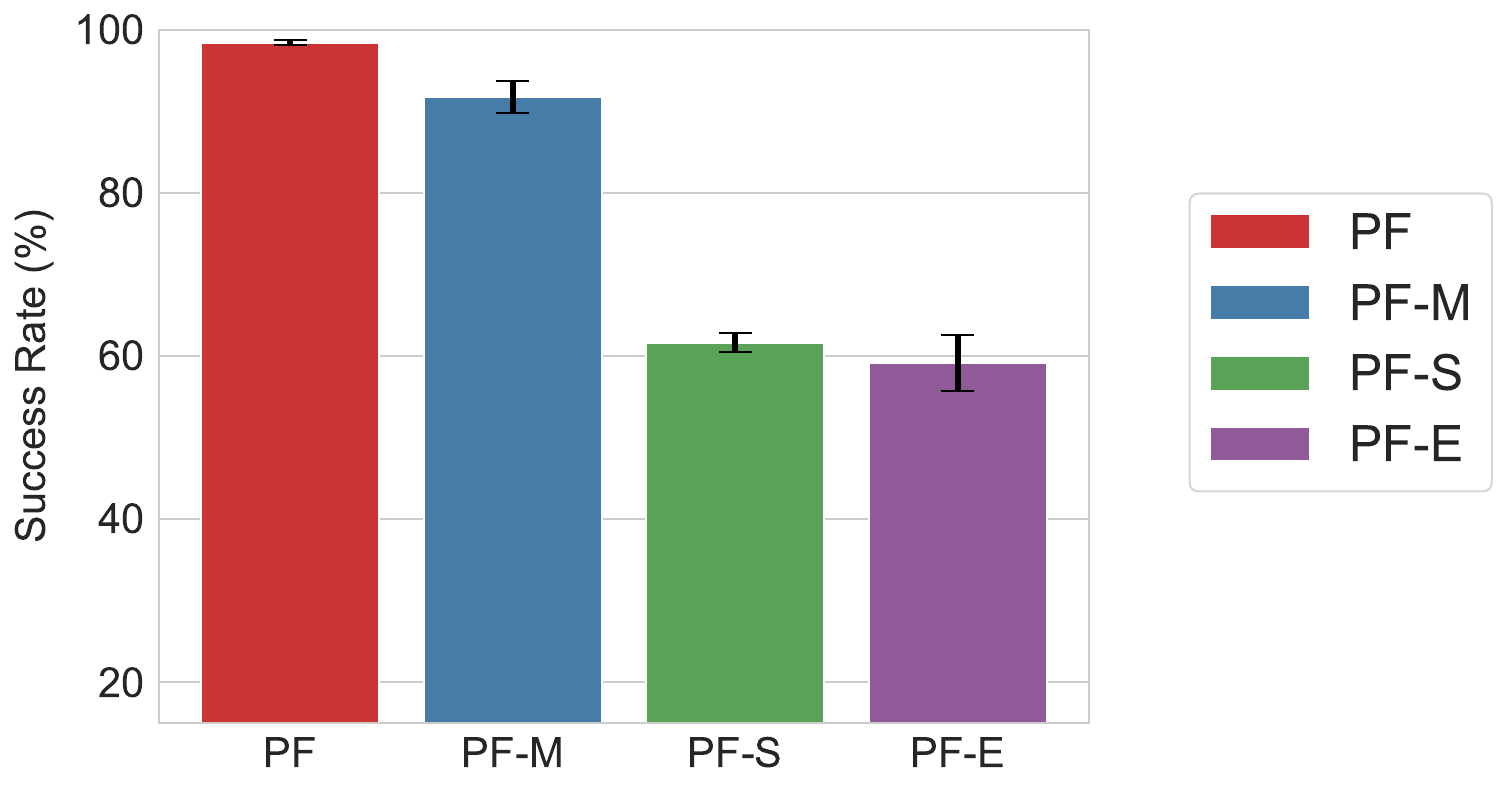}}
    \caption{The performance of our proposed Powerformer and its ablations on the IEEE 118-bus system.}
    \label{fig:ablation}
  \end{minipage}
\end{figure}

By conducting thorough ablation studies and performing a detailed analysis, we go deeper insights into our proposed Powerformer (PF) architecture. We carefully assess and quantify the contributions of each component, revealing how they individually impact the overall effectiveness and superiority of the algorithm.
Fig.~\ref{fig:ablation} presents the comparison results obtained from an array of ablation experiments conducted on IEEE 118-bus system with 10-section adjustment task.
These research findings provide valuable insights, clarifying the roles of each component in improving performance.

\noindent
\textbf{Powerformer-E (PF-E)}. This method only employs GNN to learn state representation, without performing \emph{state factorization} operation and integrating \emph{transmission section} information.
This method lacks specific feature extraction for individual electrical factors, resulting in the coupling of features. 
Moreover, the absence of section information undermines the accuracy of the representation vector in depicting the current system state, thereby impeding effective decision-making.

\noindent
\textbf{Powerformer-S (PF-S)}. This method only utilizes the representation of \texttt{Value} component $\boldsymbol{V}$ (Fig.~\ref{fig:powerformer}) as the final representation of the system state. 
In contrast to Powerformer-E, it involves the addition of an additional factorization operation. 
By analyzing the results in Fig.~\ref{fig:ablation}, it is evident that comparing Powerformer-S with Powerformer highlights the inadequacy of relying solely on state representations. 
This emphasizes the importance of incorporating section information.

\noindent
\textbf{Powerformer-M (PF-M)}. This method skips the factorization operation and use the raw four-dimensional electrical features as input to learn the final representation within the Powerformer architecture.
In contrast to Powerformer-E, 
it incorporates an additional operation for combining the transmission section information.
The less satisfactory result in Fig.~\ref{fig:ablation} implies that the disentangling operation plays a crucial role in determining the final performance.

\section{Conclusion\label{sec:conclusion}}
In this paper, we propose Powerformer, a specialized Transformer architecture designed to acquire robust state representations for the DRL agent performing the accurate action.
Specifically, our proposed Powerformer incorporates transmission section information into the existing Transformer and develops a section-adaptive attention mechanism. 
Furthermore, our approach disentangles power system graph data into independent electrical factor graphs to conduct a multi-factor attention mechanism, and employs graph neural networks to extract system state features while considering the graph topological structure of the power system.
This allows for the robust representations of the sectional power system state, enabling the development of efficient strategies to effectively manage and adjust transmission section power flow.
From the perspective of industrial applications, the utilization of our proposed architecture facilitates controlled power dispatch, leading to improved stability of the power system in real-time power flow adjustment. 

\begin{acks}
This work is supported by the Science and Technology Project of SGCC: Research and Digital Application of High-precision Electric Power Super-scale Pre-trained Visual Model (5108-202218280A-2-395-XG).
\end{acks}

\bibliographystyle{ACM-Reference-Format}
\bibliography{sample-base}


\begin{thebibliography}{47}


\ifx \showCODEN    \undefined \def \showCODEN     #1{\unskip}     \fi
\ifx \showDOI      \undefined \def \showDOI       #1{#1}\fi
\ifx \showISBNx    \undefined \def \showISBNx     #1{\unskip}     \fi
\ifx \showISBNxiii \undefined \def \showISBNxiii  #1{\unskip}     \fi
\ifx \showISSN     \undefined \def \showISSN      #1{\unskip}     \fi
\ifx \showLCCN     \undefined \def \showLCCN      #1{\unskip}     \fi
\ifx \shownote     \undefined \def \shownote      #1{#1}          \fi
\ifx \showarticletitle \undefined \def \showarticletitle #1{#1}   \fi
\ifx \showURL      \undefined \def \showURL       {\relax}        \fi
\providecommand\bibfield[2]{#2}
\providecommand\bibinfo[2]{#2}
\providecommand\natexlab[1]{#1}
\providecommand\showeprint[2][]{arXiv:#2}

\bibitem[Cao et~al\mbox{.}(2022)]%
        {cao2021deep}
\bibfield{author}{\bibinfo{person}{Di Cao}, \bibinfo{person}{Junbo Zhao}, \bibinfo{person}{Weihao Hu}, \bibinfo{person}{Nanpeng Yu}, \bibinfo{person}{Fei Ding}, \bibinfo{person}{Qi Huang}, {and} \bibinfo{person}{Zhe Chen}.} \bibinfo{year}{2022}\natexlab{}.
\newblock \showarticletitle{Deep reinforcement learning enabled physical-model-free two-timescale voltage control method for active distribution systems}.
\newblock \bibinfo{journal}{\emph{IEEE Transactions on Smart Grid}} \bibinfo{volume}{13}, \bibinfo{number}{1} (\bibinfo{year}{2022}), \bibinfo{pages}{149--165}.
\newblock


\bibitem[Capitanescu et~al\mbox{.}(2011)]%
        {capitanescu2011state}
\bibfield{author}{\bibinfo{person}{Florin Capitanescu}, \bibinfo{person}{JL~Martinez Ramos}, \bibinfo{person}{Patrick Panciatici}, \bibinfo{person}{Daniel Kirschen}, \bibinfo{person}{A~Marano Marcolini}, \bibinfo{person}{Ludovic Platbrood}, {and} \bibinfo{person}{Louis Wehenkel}.} \bibinfo{year}{2011}\natexlab{}.
\newblock \showarticletitle{State-of-the-art, challenges, and future trends in security constrained optimal power flow}.
\newblock \bibinfo{journal}{\emph{Electric power systems research}} \bibinfo{volume}{81}, \bibinfo{number}{8} (\bibinfo{year}{2011}), \bibinfo{pages}{1731--1741}.
\newblock


\bibitem[Chen et~al\mbox{.}(2023a)]%
        {chen2023improving}
\bibfield{author}{\bibinfo{person}{Kaixuan Chen}, \bibinfo{person}{Shunyu Liu}, \bibinfo{person}{Tongtian Zhu}, \bibinfo{person}{Ji Qiao}, \bibinfo{person}{Yun Su}, {et~al\mbox{.}}} \bibinfo{year}{2023}\natexlab{a}.
\newblock \showarticletitle{Improving Expressivity of GNNs with Subgraph-specific Factor Embedded Normalization}. In \bibinfo{booktitle}{\emph{Proceedings of the 29th ACM SIGKDD Conference on Knowledge Discovery and Data Mining (KDD)}}. \bibinfo{pages}{237--249}.
\newblock


\bibitem[Chen et~al\mbox{.}(2020)]%
        {chen2020covariance}
\bibfield{author}{\bibinfo{person}{Kai-Xuan Chen}, \bibinfo{person}{Jie-Yi Ren}, \bibinfo{person}{Xiao-Jun Wu}, {and} \bibinfo{person}{Josef Kittler}.} \bibinfo{year}{2020}\natexlab{}.
\newblock \showarticletitle{Covariance descriptors on a gaussian manifold and their application to image set classification}.
\newblock \bibinfo{journal}{\emph{Pattern Recognition}}  \bibinfo{volume}{107} (\bibinfo{year}{2020}), \bibinfo{pages}{107463}.
\newblock


\bibitem[Chen et~al\mbox{.}(2023c)]%
        {Chen2022RiemannianLM}
\bibfield{author}{\bibinfo{person}{Ziheng Chen}, \bibinfo{person}{Tianyang Xu}, \bibinfo{person}{Xiaojun Wu}, \bibinfo{person}{Rui Wang}, \bibinfo{person}{Zhiwu Huang}, {and} \bibinfo{person}{Josef Kittler}.} \bibinfo{year}{2023}\natexlab{c}.
\newblock \showarticletitle{Riemannian Local Mechanism for SPD Neural Networks}. In \bibinfo{booktitle}{\emph{Proceedings of the AAAI Conference on Artificial Intelligence (AAAI)}}. \bibinfo{pages}{7104--7112}.
\newblock


\bibitem[Chen et~al\mbox{.}(2023b)]%
        {chen2023hybrid}
\bibfield{author}{\bibinfo{person}{Ziheng Chen}, \bibinfo{person}{Tianyang Xu}, \bibinfo{person}{Xiao-Jun Wu}, \bibinfo{person}{Rui Wang}, {and} \bibinfo{person}{Josef Kittler}.} \bibinfo{year}{2023}\natexlab{b}.
\newblock \showarticletitle{Hybrid Riemannian Graph-Embedding Metric Learning for Image Set Classification}.
\newblock \bibinfo{journal}{\emph{IEEE Transactions on Big Data}} \bibinfo{volume}{9}, \bibinfo{number}{01} (\bibinfo{year}{2023}), \bibinfo{pages}{75--92}.
\newblock


\bibitem[Chen et~al\mbox{.}(2023d)]%
        {chen2023tele}
\bibfield{author}{\bibinfo{person}{Zhuo Chen}, \bibinfo{person}{Wen Zhang}, \bibinfo{person}{Yufeng Huang}, \bibinfo{person}{Mingyang Chen}, \bibinfo{person}{Yuxia Geng}, \bibinfo{person}{Hongtao Yu}, \bibinfo{person}{Zhen Bi}, \bibinfo{person}{Yichi Zhang}, \bibinfo{person}{Zhen Yao}, \bibinfo{person}{Wenting Song}, \bibinfo{person}{Xinliang Wu}, \bibinfo{person}{Yi Yang}, \bibinfo{person}{Mingyi Chen}, \bibinfo{person}{Zhaoyang Lian}, \bibinfo{person}{Yingying Li}, \bibinfo{person}{Lei Cheng}, {and} \bibinfo{person}{Huanjun Chen}.} \bibinfo{year}{2023}\natexlab{d}.
\newblock \showarticletitle{Tele-knowledge pre-training for fault analysis}. In \bibinfo{booktitle}{\emph{International Conference on Data Engineering (ICDE)}}. \bibinfo{pages}{3453--3466}.
\newblock


\bibitem[Dai et~al\mbox{.}(2020)]%
        {dai2019distributed}
\bibfield{author}{\bibinfo{person}{Pengcheng Dai}, \bibinfo{person}{Wenwu Yu}, \bibinfo{person}{Guanghui Wen}, {and} \bibinfo{person}{Simone Baldi}.} \bibinfo{year}{2020}\natexlab{}.
\newblock \showarticletitle{Distributed reinforcement learning algorithm for dynamic economic dispatch with unknown generation cost functions}.
\newblock \bibinfo{journal}{\emph{IEEE Transactions on Industrial Informatics}} \bibinfo{volume}{16}, \bibinfo{number}{4} (\bibinfo{year}{2020}), \bibinfo{pages}{2258--2267}.
\newblock


\bibitem[Ding et~al\mbox{.}(2024)]%
        {ding2024freecustom}
\bibfield{author}{\bibinfo{person}{Ganggui Ding}, \bibinfo{person}{Canyu Zhao}, \bibinfo{person}{Wen Wang}, \bibinfo{person}{Zhen Yang}, \bibinfo{person}{Zide Liu}, \bibinfo{person}{Hao Chen}, {and} \bibinfo{person}{Chunhua Shen}.} \bibinfo{year}{2024}\natexlab{}.
\newblock \showarticletitle{FreeCustom: Tuning-Free Customized Image Generation for Multi-Concept Composition}. In \bibinfo{booktitle}{\emph{Proceedings of the IEEE/CVF Conference on Computer Vision and Pattern Recognition}}. \bibinfo{pages}{9089--9098}.
\newblock


\bibitem[Dommel and Tinney(1968)]%
        {dommel1968optimal}
\bibfield{author}{\bibinfo{person}{Hermann~W Dommel} {and} \bibinfo{person}{William~F Tinney}.} \bibinfo{year}{1968}\natexlab{}.
\newblock \showarticletitle{Optimal power flow solutions}.
\newblock \bibinfo{journal}{\emph{IEEE Transactions on power apparatus and systems}} \bibinfo{number}{10} (\bibinfo{year}{1968}), \bibinfo{pages}{1866--1876}.
\newblock


\bibitem[Dutta and Singh(2008)]%
        {dutta2008optimal}
\bibfield{author}{\bibinfo{person}{Sudipta Dutta} {and} \bibinfo{person}{SP Singh}.} \bibinfo{year}{2008}\natexlab{}.
\newblock \showarticletitle{Optimal rescheduling of generators for congestion management based on particle swarm optimization}.
\newblock \bibinfo{journal}{\emph{IEEE transactions on Power Systems}} \bibinfo{volume}{23}, \bibinfo{number}{4} (\bibinfo{year}{2008}), \bibinfo{pages}{1560--1569}.
\newblock


\bibitem[Gao et~al\mbox{.}(2021)]%
        {gao2021hybrid}
\bibfield{author}{\bibinfo{person}{Qiu Gao}, \bibinfo{person}{Youbo Liu}, \bibinfo{person}{Junbo Zhao}, \bibinfo{person}{Junyong Liu}, {and} \bibinfo{person}{Chi~Yung Chung}.} \bibinfo{year}{2021}\natexlab{}.
\newblock \showarticletitle{Hybrid deep learning for dynamic total transfer capability control}.
\newblock \bibinfo{journal}{\emph{IEEE Transactions on Power Systems}} \bibinfo{volume}{36}, \bibinfo{number}{3} (\bibinfo{year}{2021}), \bibinfo{pages}{2733--2736}.
\newblock


\bibitem[Hasselt et~al\mbox{.}(2016)]%
        {Hasselt2015DeepRL}
\bibfield{author}{\bibinfo{person}{H.~V. Hasselt}, \bibinfo{person}{Arthur Guez}, {and} \bibinfo{person}{David Silver}.} \bibinfo{year}{2016}\natexlab{}.
\newblock \showarticletitle{Deep Reinforcement Learning with Double Q-Learning}. In \bibinfo{booktitle}{\emph{Proceedings of the AAAI Conference on Artificial Intelligence (AAAI)}}. \bibinfo{pages}{2094--2100}.
\newblock


\bibitem[Hu et~al\mbox{.}(2024)]%
        {hu2023towards}
\bibfield{author}{\bibinfo{person}{Jianxiong Hu}, \bibinfo{person}{Yujian Ye}, \bibinfo{person}{Yi Tang}, {and} \bibinfo{person}{Goran Strbac}.} \bibinfo{year}{2024}\natexlab{}.
\newblock \showarticletitle{Towards risk-aware real-time security constrained economic dispatch: A tailored deep reinforcement learning approach}.
\newblock \bibinfo{journal}{\emph{IEEE Transactions on Power Systems}} \bibinfo{volume}{39}, \bibinfo{number}{2} (\bibinfo{year}{2024}), \bibinfo{pages}{3972--3986}.
\newblock


\bibitem[Jiang et~al\mbox{.}(2022)]%
        {jiang2022steady}
\bibfield{author}{\bibinfo{person}{Yunpeng Jiang}, \bibinfo{person}{Zhouyang Ren}, \bibinfo{person}{Xin Yang}, \bibinfo{person}{Qiuyan Li}, {and} \bibinfo{person}{Yan Xu}.} \bibinfo{year}{2022}\natexlab{}.
\newblock \showarticletitle{A steady-state energy flow analysis method for integrated natural gas and power systems based on topology decoupling}.
\newblock \bibinfo{journal}{\emph{Applied Energy}}  \bibinfo{volume}{306} (\bibinfo{year}{2022}), \bibinfo{pages}{118007}.
\newblock


\bibitem[Jing(2023)]%
        {jing2023efficient}
\bibfield{author}{\bibinfo{person}{Yongcheng Jing}.} \bibinfo{year}{2023}\natexlab{}.
\newblock \emph{\bibinfo{title}{Efficient Representation Learning With Graph Neural Networks}}.
\newblock \bibinfo{thesistype}{Ph.\,D. Dissertation}.
\newblock


\bibitem[Josz et~al\mbox{.}(2016)]%
        {Josz2016ACPF}
\bibfield{author}{\bibinfo{person}{C{\'e}dric Josz}, \bibinfo{person}{St{\'e}phane Fliscounakis}, \bibinfo{person}{Jean Maeght}, {and} \bibinfo{person}{Patrick Panciatici}.} \bibinfo{year}{2016}\natexlab{}.
\newblock \showarticletitle{AC power flow data in MATPOWER and QCQP format: iTesla, RTE snapshots, and PEGASE}.
\newblock \bibinfo{journal}{\emph{arXiv preprint arXiv:1603.01533}} (\bibinfo{year}{2016}).
\newblock


\bibitem[LeCun et~al\mbox{.}(2015)]%
        {lecun2015deep}
\bibfield{author}{\bibinfo{person}{Yann LeCun}, \bibinfo{person}{Yoshua Bengio}, {and} \bibinfo{person}{Geoffrey Hinton}.} \bibinfo{year}{2015}\natexlab{}.
\newblock \showarticletitle{Deep learning}.
\newblock \bibinfo{journal}{\emph{nature}} \bibinfo{volume}{521}, \bibinfo{number}{7553} (\bibinfo{year}{2015}), \bibinfo{pages}{436--444}.
\newblock


\bibitem[Li et~al\mbox{.}(2021)]%
        {li2021virtual}
\bibfield{author}{\bibinfo{person}{Dewen Li}, \bibinfo{person}{Liying Yu}, \bibinfo{person}{Ning Li}, {and} \bibinfo{person}{Frank Lewis}.} \bibinfo{year}{2021}\natexlab{}.
\newblock \showarticletitle{Virtual-action-based coordinated reinforcement learning for distributed economic dispatch}.
\newblock \bibinfo{journal}{\emph{IEEE transactions on power systems}} \bibinfo{volume}{36}, \bibinfo{number}{6} (\bibinfo{year}{2021}), \bibinfo{pages}{5143--5152}.
\newblock


\bibitem[Li et~al\mbox{.}(2023)]%
        {Li2023MessagepassingST}
\bibfield{author}{\bibinfo{person}{Wenda Li}, \bibinfo{person}{Kaixuan Chen}, \bibinfo{person}{Shunyu Liu}, \bibinfo{person}{Wenjie Huang}, \bibinfo{person}{Haofei Zhang}, \bibinfo{person}{Yingjie Tian}, \bibinfo{person}{Yun Su}, {and} \bibinfo{person}{Mingli Song}.} \bibinfo{year}{2023}\natexlab{}.
\newblock \showarticletitle{Message-passing selection: Towards interpretable GNNs for graph classification}. In \bibinfo{booktitle}{\emph{Tiny Track @ International Conference on Learning Representations (ICLR)}}.
\newblock


\bibitem[Liu and Chu(2015)]%
        {liu2015iterative}
\bibfield{author}{\bibinfo{person}{Jian-Hong Liu} {and} \bibinfo{person}{Chia-Chi Chu}.} \bibinfo{year}{2015}\natexlab{}.
\newblock \showarticletitle{Iterative distributed algorithms for real-time available transfer capability assessment of multiarea power systems}.
\newblock \bibinfo{journal}{\emph{IEEE Transactions on Smart Grid}} \bibinfo{volume}{6}, \bibinfo{number}{5} (\bibinfo{year}{2015}), \bibinfo{pages}{2569--2578}.
\newblock


\bibitem[Liu et~al\mbox{.}(2023)]%
        {liu2023transmission}
\bibfield{author}{\bibinfo{person}{Shunyu Liu}, \bibinfo{person}{Wei Luo}, \bibinfo{person}{Yanzhen Zhou}, \bibinfo{person}{Kaixuan Chen}, \bibinfo{person}{Quan Zhang}, \bibinfo{person}{Huating Xu}, \bibinfo{person}{Qinglai Guo}, {and} \bibinfo{person}{Mingli Song}.} \bibinfo{year}{2023}\natexlab{}.
\newblock \showarticletitle{Transmission Interface Power Flow Adjustment: A Deep Reinforcement Learning Approach Based on Multi-Task Attribution Map}.
\newblock \bibinfo{journal}{\emph{IEEE Transactions on Power System}} (\bibinfo{year}{2023}).
\newblock


\bibitem[Min et~al\mbox{.}(2022)]%
        {min2022transformer}
\bibfield{author}{\bibinfo{person}{Erxue Min}, \bibinfo{person}{Runfa Chen}, \bibinfo{person}{Yatao Bian}, \bibinfo{person}{Tingyang Xu}, \bibinfo{person}{Kangfei Zhao}, \bibinfo{person}{Wenbing Huang}, \bibinfo{person}{Peilin Zhao}, \bibinfo{person}{Junzhou Huang}, \bibinfo{person}{Sophia Ananiadou}, {and} \bibinfo{person}{Yu Rong}.} \bibinfo{year}{2022}\natexlab{}.
\newblock \showarticletitle{Transformer for graphs: An overview from architecture perspective}.
\newblock \bibinfo{journal}{\emph{arXiv preprint arXiv:2202.08455}} (\bibinfo{year}{2022}).
\newblock


\bibitem[Min and Abur(2006)]%
        {min2006total}
\bibfield{author}{\bibinfo{person}{Liang Min} {and} \bibinfo{person}{Ali Abur}.} \bibinfo{year}{2006}\natexlab{}.
\newblock \showarticletitle{Total transfer capability computation for multi-area power systems}.
\newblock \bibinfo{journal}{\emph{IEEE Transactions on power systems}} \bibinfo{volume}{21}, \bibinfo{number}{3} (\bibinfo{year}{2006}), \bibinfo{pages}{1141--1147}.
\newblock


\bibitem[Mnih et~al\mbox{.}(2016)]%
        {Mnih2016AsynchronousMF}
\bibfield{author}{\bibinfo{person}{Volodymyr Mnih}, \bibinfo{person}{Adri{\`a}~Puigdom{\`e}nech Badia}, \bibinfo{person}{Mehdi Mirza}, \bibinfo{person}{Alex Graves}, \bibinfo{person}{Timothy~P. Lillicrap}, \bibinfo{person}{Tim Harley}, \bibinfo{person}{David Silver}, {and} \bibinfo{person}{Koray Kavukcuoglu}.} \bibinfo{year}{2016}\natexlab{}.
\newblock \showarticletitle{Asynchronous Methods for Deep Reinforcement Learning}. In \bibinfo{booktitle}{\emph{International Conference on Machine Learning (ICML)}}. \bibinfo{pages}{1928--1937}.
\newblock


\bibitem[Mnih et~al\mbox{.}(2015)]%
        {MnihKSRVBGRFOPB15}
\bibfield{author}{\bibinfo{person}{Volodymyr Mnih}, \bibinfo{person}{Koray Kavukcuoglu}, \bibinfo{person}{David Silver}, \bibinfo{person}{Andrei~A. Rusu}, \bibinfo{person}{Joel Veness}, \bibinfo{person}{Marc~G. Bellemare}, \bibinfo{person}{Alex Graves}, \bibinfo{person}{Martin~A. Riedmiller}, \bibinfo{person}{Andreas~Kirkeby Fidjeland}, {et~al\mbox{.}}} \bibinfo{year}{2015}\natexlab{}.
\newblock \showarticletitle{Human-level control through deep reinforcement learning}.
\newblock \bibinfo{journal}{\emph{Nature}} \bibinfo{volume}{518}, \bibinfo{number}{7540} (\bibinfo{year}{2015}), \bibinfo{pages}{529--533}.
\newblock


\bibitem[Mu et~al\mbox{.}(2024)]%
        {mu2023graph}
\bibfield{author}{\bibinfo{person}{Chaoxu Mu}, \bibinfo{person}{Zhaoyang Liu}, \bibinfo{person}{Jun Yan}, \bibinfo{person}{Hongjie Jia}, {and} \bibinfo{person}{Xiaoyu Zhang}.} \bibinfo{year}{2024}\natexlab{}.
\newblock \showarticletitle{Graph multi-agent reinforcement learning for inverter-based active voltage control}.
\newblock \bibinfo{journal}{\emph{IEEE Transactions on Smart Grid}} \bibinfo{volume}{15}, \bibinfo{number}{2} (\bibinfo{year}{2024}), \bibinfo{pages}{1399--1409}.
\newblock


\bibitem[Qing et~al\mbox{.}(2024)]%
        {qinga2po}
\bibfield{author}{\bibinfo{person}{Yunpeng Qing}, \bibinfo{person}{Shunyu Liu}, \bibinfo{person}{Jingyuan Cong}, \bibinfo{person}{Kaixuan Chen}, \bibinfo{person}{Yihe Zhou}, {and} \bibinfo{person}{Mingli Song}.} \bibinfo{year}{2024}\natexlab{}.
\newblock \showarticletitle{A2PO: Towards Effective Offline Reinforcement Learning from an Advantage-aware Perspective}. In \bibinfo{booktitle}{\emph{The Thirty-eighth Annual Conference on Neural Information Processing Systems}}.
\newblock


\bibitem[Schulman et~al\mbox{.}(2017)]%
        {schulman2017proximal}
\bibfield{author}{\bibinfo{person}{John Schulman}, \bibinfo{person}{Filip Wolski}, \bibinfo{person}{Prafulla Dhariwal}, \bibinfo{person}{Alec Radford}, {and} \bibinfo{person}{Oleg Klimov}.} \bibinfo{year}{2017}\natexlab{}.
\newblock \showarticletitle{Proximal policy optimization algorithms}.
\newblock \bibinfo{journal}{\emph{arXiv preprint arXiv:1707.06347}} (\bibinfo{year}{2017}).
\newblock


\bibitem[Shi et~al\mbox{.}(2024)]%
        {shi2024scene}
\bibfield{author}{\bibinfo{person}{Xuehuai Shi}, \bibinfo{person}{Lili Wang}, \bibinfo{person}{Xinda Liu}, \bibinfo{person}{Jian Wu}, {and} \bibinfo{person}{Zhiwen Shao}.} \bibinfo{year}{2024}\natexlab{}.
\newblock \showarticletitle{Scene-aware Foveated Neural Radiance Fields}.
\newblock \bibinfo{journal}{\emph{IEEE Transactions on Visualization and Computer Graphics}} (\bibinfo{year}{2024}).
\newblock


\bibitem[Shi et~al\mbox{.}(2021)]%
        {shi2021foveated}
\bibfield{author}{\bibinfo{person}{Xuehuai Shi}, \bibinfo{person}{Lili Wang}, \bibinfo{person}{Xiaoheng Wei}, {and} \bibinfo{person}{Ling-Qi Yan}.} \bibinfo{year}{2021}\natexlab{}.
\newblock \showarticletitle{Foveated photon mapping}.
\newblock \bibinfo{journal}{\emph{IEEE Transactions on Visualization and Computer Graphics}} \bibinfo{volume}{27}, \bibinfo{number}{11} (\bibinfo{year}{2021}), \bibinfo{pages}{4183--4193}.
\newblock


\bibitem[Shi et~al\mbox{.}(2022)]%
        {shi2022foveated}
\bibfield{author}{\bibinfo{person}{Xuehuai Shi}, \bibinfo{person}{Lili Wang}, \bibinfo{person}{Jian Wu}, \bibinfo{person}{Runze Fan}, {and} \bibinfo{person}{Aimin Hao}.} \bibinfo{year}{2022}\natexlab{}.
\newblock \showarticletitle{Foveated Stochastic Lightcuts}.
\newblock \bibinfo{journal}{\emph{IEEE Transactions on Visualization and Computer Graphics}} \bibinfo{volume}{28}, \bibinfo{number}{11} (\bibinfo{year}{2022}), \bibinfo{pages}{3684--3693}.
\newblock


\bibitem[Vaswani et~al\mbox{.}(2017)]%
        {vaswani2017attention}
\bibfield{author}{\bibinfo{person}{Ashish Vaswani}, \bibinfo{person}{Noam Shazeer}, \bibinfo{person}{Niki Parmar}, \bibinfo{person}{Jakob Uszkoreit}, \bibinfo{person}{Llion Jones}, \bibinfo{person}{Aidan~N Gomez}, \bibinfo{person}{{\L}ukasz Kaiser}, {and} \bibinfo{person}{Illia Polosukhin}.} \bibinfo{year}{2017}\natexlab{}.
\newblock \showarticletitle{Attention is all you need}. In \bibinfo{booktitle}{\emph{Neural Information Processing Systems (NeurIPS)}}. \bibinfo{pages}{6000--6010}.
\newblock


\bibitem[Wang et~al\mbox{.}(2022)]%
        {wang2022stabilizing}
\bibfield{author}{\bibinfo{person}{Minrui Wang}, \bibinfo{person}{Mingxiao Feng}, \bibinfo{person}{Wengang Zhou}, {and} \bibinfo{person}{Houqiang Li}.} \bibinfo{year}{2022}\natexlab{}.
\newblock \showarticletitle{Stabilizing voltage in power distribution networks via multi-agent reinforcement learning with transformer}. In \bibinfo{booktitle}{\emph{Proceedings of the 28th ACM SIGKDD Conference on Knowledge Discovery and Data Mining (KDD)}}. \bibinfo{pages}{1899--1909}.
\newblock


\bibitem[Wang et~al\mbox{.}(2016)]%
        {Wang2015DuelingNA}
\bibfield{author}{\bibinfo{person}{Ziyu Wang}, \bibinfo{person}{Tom Schaul}, \bibinfo{person}{Matteo Hessel}, \bibinfo{person}{Hado Hasselt}, \bibinfo{person}{Marc Lanctot}, {and} \bibinfo{person}{Nando Freitas}.} \bibinfo{year}{2016}\natexlab{}.
\newblock \showarticletitle{Dueling Network Architectures for Deep Reinforcement Learning}. In \bibinfo{booktitle}{\emph{International Conference on Machine Learning (ICML)}}. \bibinfo{pages}{1995--2003}.
\newblock


\bibitem[Wei et~al\mbox{.}(2023)]%
        {wei2023agent}
\bibfield{author}{\bibinfo{person}{Yaoquan Wei}, \bibinfo{person}{Shunyu Liu}, \bibinfo{person}{Jie Song}, \bibinfo{person}{Tongya Zheng}, \bibinfo{person}{Kaixuan Chen}, \bibinfo{person}{Yong Wang}, {and} \bibinfo{person}{Mingli Song}.} \bibinfo{year}{2023}\natexlab{}.
\newblock \showarticletitle{Agent-Aware Training for Agent-Agnostic Action Advising in Deep Reinforcement Learning}.
\newblock \bibinfo{journal}{\emph{arXiv preprint arXiv:2311.16807}} (\bibinfo{year}{2023}).
\newblock


\bibitem[Wu et~al\mbox{.}(2024)]%
        {wu2023constrained}
\bibfield{author}{\bibinfo{person}{Tong Wu}, \bibinfo{person}{Anna Scaglione}, {and} \bibinfo{person}{Daniel Arnold}.} \bibinfo{year}{2024}\natexlab{}.
\newblock \showarticletitle{Constrained reinforcement learning for predictive control in real-time stochastic dynamic optimal power flow}.
\newblock \bibinfo{journal}{\emph{IEEE Transactions on Power Systems}} \bibinfo{volume}{39}, \bibinfo{number}{3} (\bibinfo{year}{2024}), \bibinfo{pages}{5077--5090}.
\newblock


\bibitem[Xu et~al\mbox{.}(2019)]%
        {xu2018powerful}
\bibfield{author}{\bibinfo{person}{Keyulu Xu}, \bibinfo{person}{Weihua Hu}, \bibinfo{person}{Jure Leskovec}, {and} \bibinfo{person}{Stefanie Jegelka}.} \bibinfo{year}{2019}\natexlab{}.
\newblock \showarticletitle{How Powerful are Graph Neural Networks?}. In \bibinfo{booktitle}{\emph{International Conference on Learning Representations (ICLR)}}.
\newblock


\bibitem[Yang et~al\mbox{.}(2021)]%
        {yang2021graphformers}
\bibfield{author}{\bibinfo{person}{Junhan Yang}, \bibinfo{person}{Zheng Liu}, \bibinfo{person}{Shitao Xiao}, \bibinfo{person}{Chaozhuo Li}, \bibinfo{person}{Defu Lian}, \bibinfo{person}{Sanjay Agrawal}, \bibinfo{person}{Amit Singh}, \bibinfo{person}{Guangzhong Sun}, {and} \bibinfo{person}{Xing Xie}.} \bibinfo{year}{2021}\natexlab{}.
\newblock \showarticletitle{Graphformers: Gnn-nested transformers for representation learning on textual graph}.
\newblock \bibinfo{journal}{\emph{Advances in Neural Information Processing Systems}}  \bibinfo{volume}{34} (\bibinfo{year}{2021}), \bibinfo{pages}{28798--28810}.
\newblock


\bibitem[Yang et~al\mbox{.}(2020)]%
        {yang2019two}
\bibfield{author}{\bibinfo{person}{Qiuling Yang}, \bibinfo{person}{Gang Wang}, \bibinfo{person}{Alireza Sadeghi}, \bibinfo{person}{Georgios~B Giannakis}, {and} \bibinfo{person}{Jian Sun}.} \bibinfo{year}{2020}\natexlab{}.
\newblock \showarticletitle{Two-timescale voltage control in distribution grids using deep reinforcement learning}.
\newblock \bibinfo{journal}{\emph{IEEE Transactions on Smart Grid}} \bibinfo{volume}{11}, \bibinfo{number}{3} (\bibinfo{year}{2020}), \bibinfo{pages}{2313--2323}.
\newblock


\bibitem[Ying et~al\mbox{.}(2021)]%
        {ying2021transformers}
\bibfield{author}{\bibinfo{person}{Chengxuan Ying}, \bibinfo{person}{Tianle Cai}, \bibinfo{person}{Shengjie Luo}, \bibinfo{person}{Shuxin Zheng}, \bibinfo{person}{Guolin Ke}, \bibinfo{person}{Di He}, \bibinfo{person}{Yanming Shen}, {and} \bibinfo{person}{Tie-Yan Liu}.} \bibinfo{year}{2021}\natexlab{}.
\newblock \showarticletitle{Do transformers really perform badly for graph representation?}
\newblock \bibinfo{journal}{\emph{Advances in neural information processing systems}}  \bibinfo{volume}{34} (\bibinfo{year}{2021}), \bibinfo{pages}{28877--28888}.
\newblock


\bibitem[Yu et~al\mbox{.}(2024)]%
        {yu2024multi}
\bibfield{author}{\bibinfo{person}{Na Yu}, \bibinfo{person}{Ke Xu}, \bibinfo{person}{Kaixuan Chen}, \bibinfo{person}{Shunyu Liu}, \bibinfo{person}{Tongya Zheng}, {and} \bibinfo{person}{Mingli Song}.} \bibinfo{year}{2024}\natexlab{}.
\newblock \showarticletitle{Multi-Channel Graph Fusion Representation for Tabular Data Imputation}. In \bibinfo{booktitle}{\emph{2024 International Joint Conference on Neural Networks (IJCNN)}}. IEEE, \bibinfo{pages}{1--8}.
\newblock


\bibitem[Yun et~al\mbox{.}(2019)]%
        {yun2019graph}
\bibfield{author}{\bibinfo{person}{Seongjun Yun}, \bibinfo{person}{Minbyul Jeong}, \bibinfo{person}{Raehyun Kim}, \bibinfo{person}{Jaewoo Kang}, {and} \bibinfo{person}{Hyunwoo~J Kim}.} \bibinfo{year}{2019}\natexlab{}.
\newblock \showarticletitle{Graph transformer networks}.
\newblock \bibinfo{journal}{\emph{Advances in neural information processing systems}}  \bibinfo{volume}{32} (\bibinfo{year}{2019}).
\newblock


\bibitem[Zhang et~al\mbox{.}(2022)]%
        {zhang2022bootstrapping}
\bibfield{author}{\bibinfo{person}{Haofei Zhang}, \bibinfo{person}{Jiarui Duan}, \bibinfo{person}{Mengqi Xue}, \bibinfo{person}{Jie Song}, \bibinfo{person}{Li Sun}, {and} \bibinfo{person}{Mingli Song}.} \bibinfo{year}{2022}\natexlab{}.
\newblock \showarticletitle{Bootstrapping ViTs: Towards liberating vision transformers from pre-training}. In \bibinfo{booktitle}{\emph{Proceedings of the IEEE/CVF Conference on Computer Vision and Pattern Recognition}}. \bibinfo{pages}{8944--8953}.
\newblock


\bibitem[Zhang and Yang(2018)]%
        {zhang2018overview}
\bibfield{author}{\bibinfo{person}{Yu Zhang} {and} \bibinfo{person}{Qiang Yang}.} \bibinfo{year}{2018}\natexlab{}.
\newblock \showarticletitle{An overview of multi-task learning}.
\newblock \bibinfo{journal}{\emph{National Science Review}} \bibinfo{volume}{5}, \bibinfo{number}{1} (\bibinfo{year}{2018}), \bibinfo{pages}{30--43}.
\newblock


\bibitem[Zhou and Yang(2023)]%
        {Zhou2023AutomaticTR}
\bibfield{author}{\bibinfo{person}{Menghui Zhou} {and} \bibinfo{person}{Po Yang}.} \bibinfo{year}{2023}\natexlab{}.
\newblock \showarticletitle{Automatic Temporal Relation in Multi-Task Learning}. In \bibinfo{booktitle}{\emph{Proceedings of the 29th ACM SIGKDD Conference on Knowledge Discovery and Data Mining (KDD)}}. \bibinfo{pages}{3570--3580}.
\newblock


\bibitem[Zhu et~al\mbox{.}(2023)]%
        {Zhu2023OnSE}
\bibfield{author}{\bibinfo{person}{Wenhao Zhu}, \bibinfo{person}{Tianyu Wen}, \bibinfo{person}{Guojie Song}, \bibinfo{person}{Liangji Wang}, {and} \bibinfo{person}{Bo Zheng}.} \bibinfo{year}{2023}\natexlab{}.
\newblock \showarticletitle{On Structural Expressive Power of Graph Transformers}. In \bibinfo{booktitle}{\emph{Proceedings of the 29th ACM SIGKDD Conference on Knowledge Discovery and Data Mining (KDD)}}. \bibinfo{pages}{3628--3637}.
\newblock


\end{thebibliography}

\newpage
\appendix

\begin{table}
  \centering
  \normalsize
  \renewcommand{\arraystretch}{0.9} 
  \caption{The pre-scheduled lower and upper power bound (MW) of each transmission section in the IEEE 118-bus system, the realistic China 300-bus system, and the European 9241-bus system.}
  \label{tab:range_supp}
  \resizebox{0.45\textwidth}{!}{%
  \begin{tabular}{@{}ccccccc@{}}
  \toprule
  \multirow{2}{*}{ \textbf{Section $\Phi$}} & \multicolumn{2}{c}{\textbf{118-bus System}} & \multicolumn{2}{c}{\textbf{300-bus System}} & \multicolumn{2}{c}{\textbf{9241-bus System}} \\ \cmidrule(l){2-7} 
  \multicolumn{1}{c}{} & \multicolumn{1}{c}{\textbf{Lower}} & \multicolumn{1}{c}{\textbf{Upper}} & \multicolumn{1}{c}{\textbf{Lower}} & \multicolumn{1}{c}{\textbf{Upper}} & \multicolumn{1}{c}{\textbf{Lower}} & \multicolumn{1}{c}{\textbf{Upper}} \\ \midrule
  1 & 90 & 640 & 140 & 1000 & 15 & 110 \\ 
  2 & 50 & 360 & 280 & 1960 & 840 & 5880 \\ 
  3 & 40 & 290 & 170 & 1200 & 145 & 1000 \\ 
  4 & 90 & 640 & 240 & 1680 & 55 & 390 \\ 
  5 & 70 & 480 & 460 & 3200 & 560 & 3920 \\ 
  6 & 45 & 300 & 200 & 1400 & 360 & 2520 \\ 
  7 & 130 & 880 & 200 & 1400 & 345 & 2410 \\ 
  8 & 55 & 390 & 480 & 3360 & 760 & 5320 \\ 
  9 & 130 & 880 & 170 & 1200 & 130 & 910 \\ 
  10 & 90 & 615 & 80 & 590 & 180 & 1260 \\  \bottomrule
  \end{tabular}%
  }
\vspace{-0.5cm}
\end{table}

\begin{figure*}
  \centering
  \begin{minipage}{1.0\linewidth}
    \centering
    \subfigure{\includegraphics[scale=0.24]{figures/legend_nosoft.pdf}}
    \vspace{-0.1cm}
    \\
    \addtocounter{subfigure}{-1}

    \subfigure[IEEE 118-bus System (4-section Task)]{\includegraphics[width=0.31\textwidth]{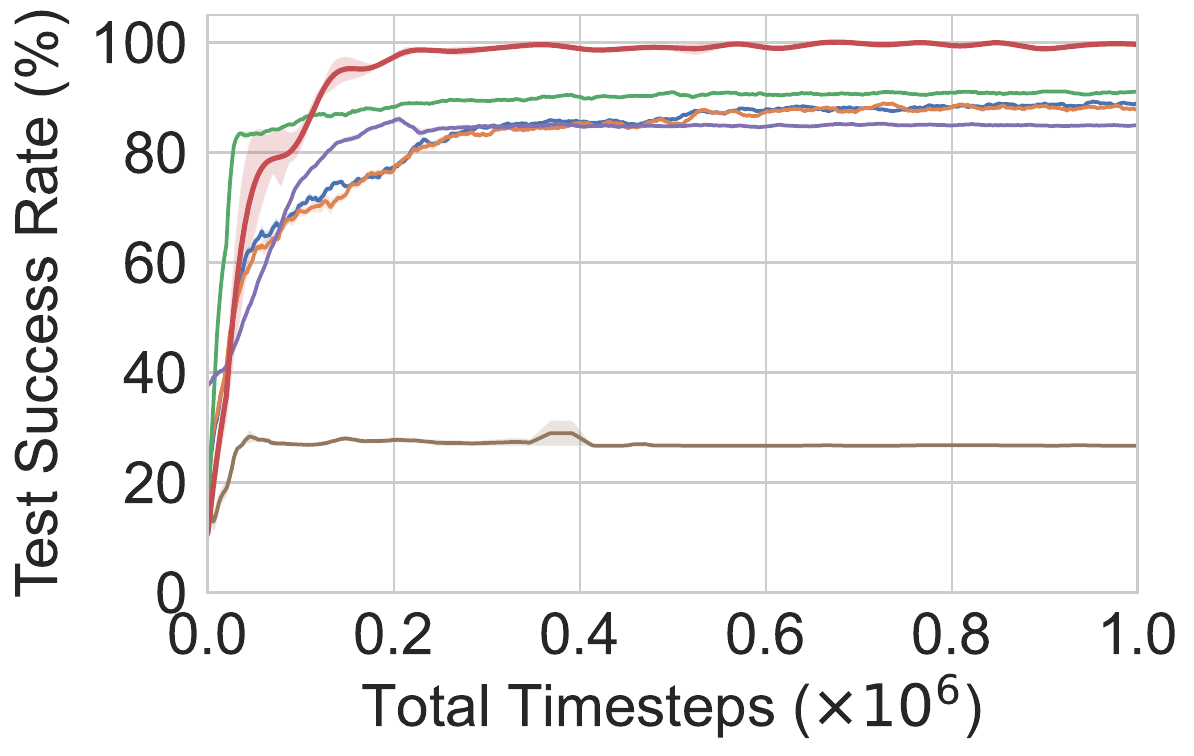}\label{fig:task4-118_supp}}\hfil
    \subfigure[Realistic 300-bus System (4-section Task)]{\includegraphics[width=0.31\textwidth]{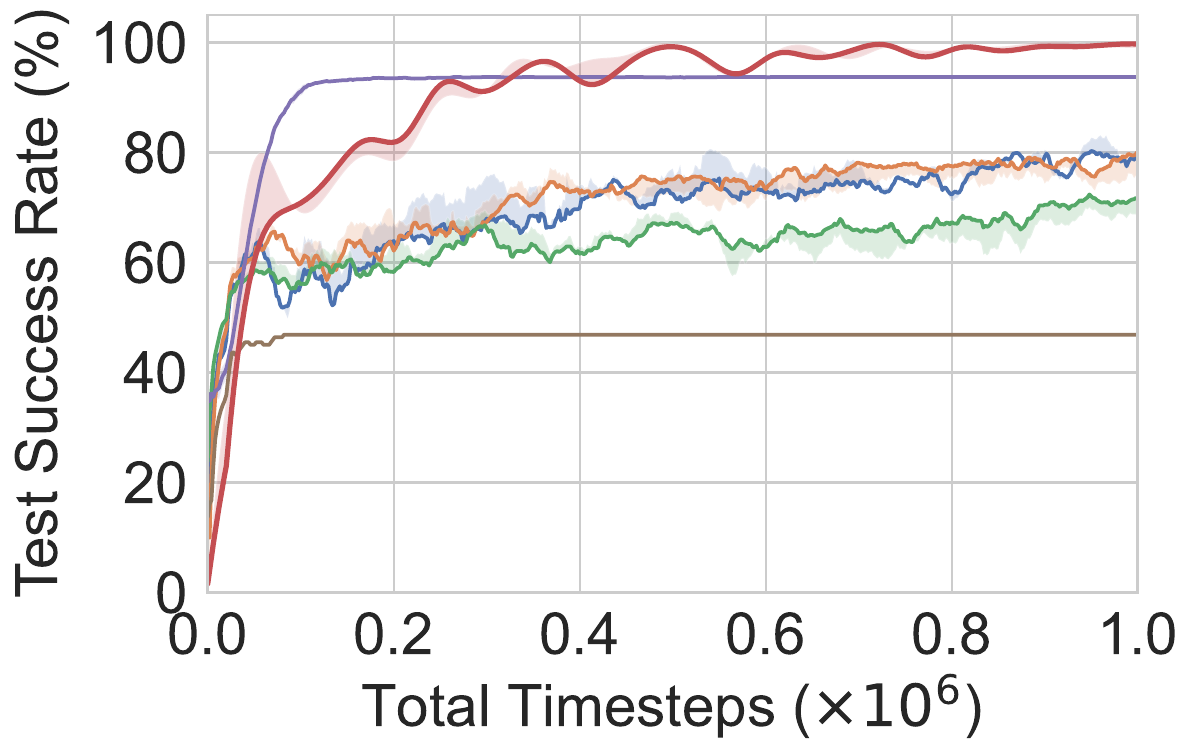}\label{fig:task4-300_supp}} \hfil
    \subfigure [European 9241-bus System (4-section Task)]{\includegraphics[width=0.3\textwidth]{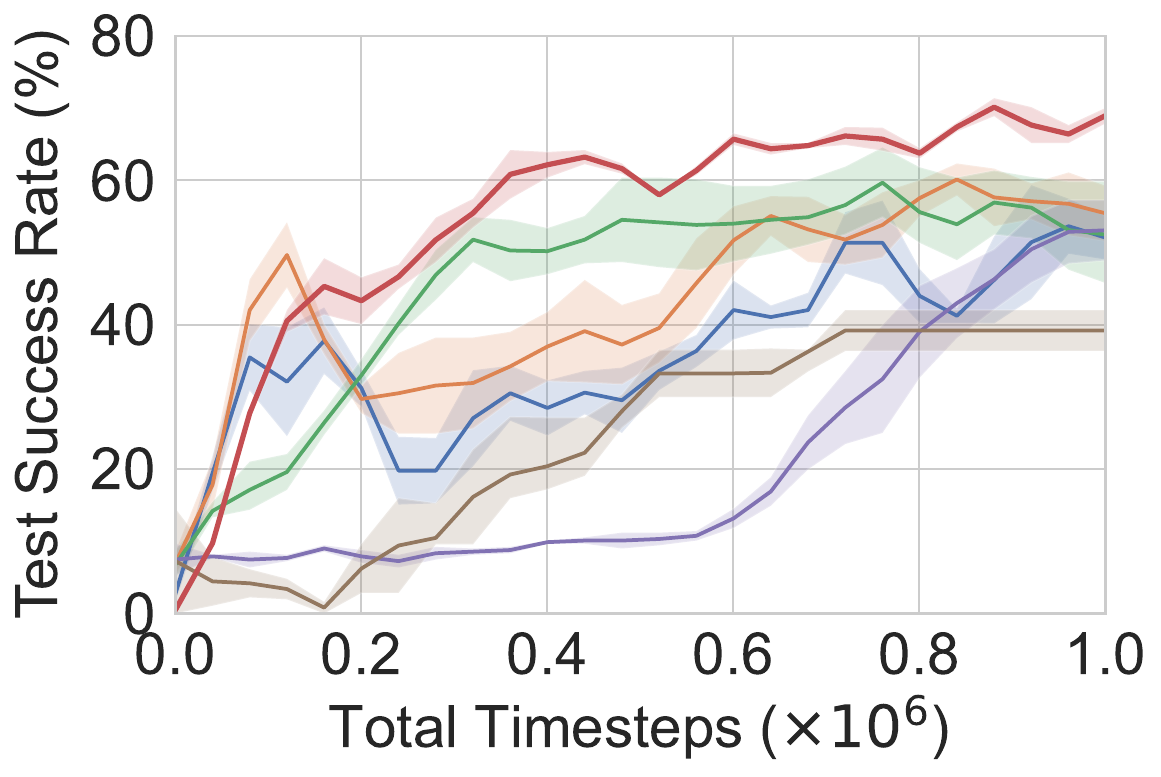}\label{fig:task4-9241_supp}} \\
    \vspace{-0.3cm}
    \subfigure[IEEE 118-bus System (10-section Task)]{\includegraphics[width=0.31\textwidth]{figures/S10case118_success_smooth_nosoft.pdf}\label{fig:task10-118_supp}} \hfil
    \subfigure[Realistic 300-bus System (10-section Task)]{\includegraphics[width=0.31\textwidth]{figures/S10case300_success_smooth_nosoft.pdf}\label{fig:task10-300_supp}} \hfil
    \subfigure[European 9241-bus System (10-section Task)]{\includegraphics[width=0.3\textwidth]{figures/S10case9241_success_smooth_nosoft.pdf}\label{fig:task10-9241_supp}}
    \caption{The learning curves of all methods for 4-section task and 10-section task on three power systems. The experimental results use the median performance and one standard deviation (shaded region) over 5 random seeds to ensure a fair comparison.}
    \label{fig:learning_curve_supp}
  \end{minipage}
  \\\vspace{0.2cm}
  \begin{minipage}{1.0\linewidth}
    \centering
    \normalsize    
  \renewcommand{\arraystretch}{1.0} 
    \captionof{table}{{The performance of our method and baselines in 4-section and 10-section tasks on three different power systems is evaluated based on an average of 5 trials. The performance differences between our method and baselines are indicated in parentheses, with better performance denoted by a higher test success rate and lower test economic cost.}}
    \label{tab:total_supp}
    \resizebox{0.98\textwidth}{!}{%
    \begin{tabular}{@{}lcccccc@{}}
      \toprule
      \multicolumn{1}{l}{\multirow{2}{*}{{Method}}} 
      & \multicolumn{2}{c}{{118-bus System~(4-section Task)}} 
      & \multicolumn{2}{c}{{300-bus System~(4-section Task)}} 
      & \multicolumn{2}{c}{{9241-bus System~(4-section Task)}} 
      \\ \cmidrule(l){2-3} \cmidrule(l){4-5} \cmidrule(l){6-7} 
      & \multicolumn{1}{c}{{Success Rate (\%)}} & \multicolumn{1}{c}{{Economic Cost (\$)}} 
      & \multicolumn{1}{c}{{Success Rate (\%)}} & \multicolumn{1}{c}{{Economic Cost (\$)}}  
      & \multicolumn{1}{c}{{Success Rate (\%)}} & \multicolumn{1}{c}{{Economic Cost (\$)}} \\ \midrule
      DQN           
      & 88.97{\normalsize{~(-19.66)}} & 693,272{\;\normalsize{~(+1,130)}} & 79.05{\normalsize{~(-20.68)}} 
      & 981,695{\normalsize{~(+19,605)}} & 52.04{\normalsize{~(-16.93)}} &  325,331{\;\;\;\;\normalsize{~(-99)}}          \\  \specialrule{0em}{1pt}{1pt}
      Double DQN        
      & 87.63{\normalsize{~(-12.04)}} & 693,653{\;\normalsize{~(+1,511)}} & 80.15{\normalsize{~(-19.58)}} 
      & 995,038{\normalsize{~(+32,948)}} & 55.41{\normalsize{~(-13.56)}} & 325,464{\;\;\;\;\normalsize{~(+34)}}             \\ \specialrule{0em}{1pt}{1pt}
      Dueling DQN 
      & 91.17{\;\normalsize{~(-8.50)}} & 692,378{\;\;\;\normalsize{~(+236)}} & 71.45{\normalsize{~(-28.28)}} 
      & 995,279{\normalsize{~(+33,189)}} & 52.48{\normalsize{~(-16.49)}} & 325,266{\;\;\;\normalsize{~(-164)}}            \\ \specialrule{0em}{1pt}{1pt}
      A2C               
      & 84.98{\normalsize{~(-14.69)}} & 697,853{\;\normalsize{~(+5,711)}} & 93.71{\;\normalsize{~(-6.02)}} 
      & 974,455{\normalsize{~(+12,365)}} & 53.07{\normalsize{~(-15.90)}} & 325,093{\;\;\;\normalsize{~(-337)}}          \\ \specialrule{0em}{1pt}{1pt}
      PPO              
      & 26.67{\normalsize{~(-73.00)}} & 673,399{\normalsize{~(-18,743)}} & 46.88{\normalsize{~(-52.85)}} 
      & 937,454{\normalsize{~(-24,636)}} & 39.18{\normalsize{~(-29.79)}} & 326,614{\;\normalsize{~(+1,184)}}            \\ \specialrule{0em}{1pt}{1pt}
      OPF                
      & 83.33{\normalsize{~(-16.34)}} & 639,782{\normalsize{~(-52,360)}} & 56.78{\normalsize{~(-42.59)}} 
      & 956,120{\;\normalsize{~(-5,970)}} & 64.47{\;\normalsize{~(-4.50)}} & 324,995{\;\;\;\normalsize{~(-435)}}         \\ \midrule
      \textbf{Powerformer~(Ours)}       
      & \textbf{99.67} & \textbf{692,142} & \textbf{99.73} 
      & \textbf{962,090} & \textbf{68.97} & \textbf{325,430}           \\ \specialrule{0em}{1pt}{1pt} \toprule
      \multicolumn{1}{l}{\multirow{2}{*}{{Method}}} 
      & \multicolumn{2}{c}{{118-bus System~(10-section Task)}} 
      & \multicolumn{2}{c}{{300-bus System~(10-section Task)}} 
      & \multicolumn{2}{c}{{9241-bus System~(10-section Task)}} 
      \\ \cmidrule(l){2-3} \cmidrule(l){4-5} \cmidrule(l){6-7} 
      & \multicolumn{1}{c}{{Success Rate (\%)}} & \multicolumn{1}{c}{{Economic Cost (\$)}} 
      & \multicolumn{1}{c}{{Success Rate (\%)}} & \multicolumn{1}{c}{{Economic Cost (\$)}}  
      & \multicolumn{1}{c}{{Success Rate (\%)}} & \multicolumn{1}{c}{{Economic Cost (\$)}} \\ \midrule
      DQN          
      & 61.23{\normalsize{~(-36.96)}}     & 626,808{\;\normalsize{~(+4,610)}} & 56.74{\normalsize{~(-40.62)}} 
      & 991,762{\normalsize{~(+22,080)}}  & 13.67{\normalsize{~(-54.58)}}     & 324,658{\normalsize{\;\;\;~(+164)}}           \\  \specialrule{0em}{1pt}{1pt}
      Double DQN        
      & 53.28{\normalsize{~(-44.91)}}     & 638,079{\normalsize{~(+15,881)}}  & 61.24{\normalsize{~(-36.12)}} 
      & 990,126{\normalsize{~(+20,444)}}  & 16.75{\normalsize{~(-42.14)}}     & 324,792{\normalsize{\;\;\;~(+305)}}           \\ \specialrule{0em}{1pt}{1pt}
      Dueling DQN      
      & 68.44{\normalsize{~(-29.75)}}     & 626,215{\;\normalsize{~(+4,017)}} & 57.66{\normalsize{~(-39.70)}}
      & 992,048{\normalsize{~(+22,366)}}  & 37.57{\normalsize{~(-30.68)}}    & 324,485{\normalsize{\;\;\;\;\;\;\;~(-2)}}            \\ \specialrule{0em}{1pt}{1pt}
      A2C                
      & 80.14{\normalsize{~(-18.05)}}     & 626,703{\;\normalsize{~(+4,505)}} & 75.61{\normalsize{~(-21.75)}}
      & 976,467{\;\normalsize{~(+6,785)}} & 12.70{\normalsize{~(-55.55)}}     & 325,000{\normalsize{\;\;\;~(+513)}}           \\ \specialrule{0em}{1pt}{1pt}
      PPO               
      & 24.28{\normalsize{~(-73.19)}}     & 596,315{\normalsize{~(-25,883)}}  & 48.61{\normalsize{~(-48.75)}} 
      & 930,581{\normalsize{~(-39,101)}}  & 19.40{\normalsize{~(-48.85)}}     & 325,549{\normalsize{~(+1,062)}}          \\ \specialrule{0em}{1pt}{1pt}
      {OPF}                        
      & 72.25{\normalsize{~(-25.94)}}     & 606,259{\normalsize{~(-15,939)}}  & 50.56{\normalsize{~(-46.80)}} 
      & 953,349{\normalsize{~(-16,333)}}  & 51.32{\normalsize{~(-16.93)}}    & {321,586{\normalsize{~(-2,901)}}}              \\ \midrule
      \textbf{Powerformer~(Ours)}              
      & \textbf{98.19}              & {\textbf{622,198}}            & \textbf{97.36} 
      & {\textbf{969,682}}            & \textbf{68.25}              & {\textbf{324,487}}          \\ \bottomrule
      \end{tabular}%
  }
\end{minipage}
\end{figure*}

\section{Scenario Generation}\label{Scenario_Generation_supp}
To obtain adequate scenario, we adopt the following steps:

\begin{itemize}[leftmargin=*]
  \item To simulate a variety of scenarios within the power system more effectively, we implement random perturbations to loads and generators. For each scenario, we randomly select 25\% of generators and loads and then introduce random disturbances to their initial power flow ranging from 10\% to 200\%, with $10\%$ intervals.

  \item The power range for each transmission section $\Phi$ is pre-scheduled, as detailed in Table~\ref{tab:range_supp}. Insecure scenarios, where the power flow of the corresponding transmission sections exceeds pre-scheduled power range, are considered for case studies during each power flow adjustment.
\end{itemize}

Finally, a total of 1829 scenarios were generated for the IEEE 118-bus system, with 1656 scenarios used for training and 173 for testing. Similarly, the China 300-bus system resulted in 1817 scenarios, including 1637 for training and 180 for testing. The European 9241-bus system produced 2037 scenarios, with 1848 for training and 189 for testing. 
The generated scenario cases indicate that the section flow information has exceeded the predefined range. Our goal is to adjust the section flow within the predetermined range through the power generation of generators.

\section{Parameter setting}\label{parameter_setting_supp}

The proposed Powerformer employs a two-layer networks configuration, specifically the hidden dimension $d$ is 64. 
Additionally, the section encoder utilizes a two-layer MLP network with a candidate hidden dimension set \{128, 64\}.
To implemente the Powerformer, we utilize the Before Transformer architecture~\cite{min2022transformer}, a type of graph transformer.
The Dueling architecture includes a two-layer value network with a dimension of 128 and a one-layer advantage network with a dimension of 128. The section representation is
$\boldsymbol{Z}_\Phi \in \mathbb{R}^{4m}$, where 4 is the corresponding electrical factors of active power, reactive power, voltage magnitude, and phase angle, $m$ denotes the number of the transmission lines.
All network modules in the architecture employ the ReLU activation function. During each training iteration, batches of 64 episodes are sampled from a replay buffer containing 20,000 episodes. The target update interval is set to 100, and the discount factor is 0.9. For exploration, we apply an epsilon-greedy approach, where epsilon linearly decreases from 0.1 to 0.01 over 500,000 training steps and remains constant thereafter.

\section{Comparison for both 4-section and 10-section adjustment task}\label{total_section_task_supp}

To demonstrate the superiority of our method, we thoroughly analyze and compare our method with various popular approaches in DRL framework, as well as a well-established traditional optimal power flow (OPF) adjustment method. Moreover, in order to thoroughly validate the effectiveness of our Powerformer, we design the additional 4-section power flow adjustment tasks, which aims to deal with the top four ranked sections in terms of transmission capacity among a total of ten sections.

Through analyzing the learning curve in Fig.~\ref{fig:learning_curve_supp}, 
our proposed method has successfully achieved the best test success rate among all considered architectures on three different systems. 
We can analyze and compare the superiority of our method based on the difficulty levels of different tasks, especially in the more challenging task of 10-section adjustment (Fig.~\ref{fig:task10-118_supp},~\ref{fig:task10-300_supp} and~\ref{fig:task10-9241_supp}), where the effectiveness of our method is highlighted.
Furthermore, from the perspective of different system cases, our method performs better on large-scale datasets. 
This is because our method is able to learn better representations of sectional power system states, thus avoiding the coupling of state features. 
We can comparatively analyze the learning curve on the IEEE 118-bus and European 9241-bus systems, and the significant differences in the effectiveness of our method can be directly observed. 

From a quantifiable perspective of comparison adjustment algorithms, we leverage the comparative results of success rates and economic costs pertaining to 4-section and 10-section adjustment tasks, as evidenced in Table~\ref{tab:total_supp}. Specifically, we further analyze the results by examining different power system cases. 
For the IEEE 118-bus system, 
Powerformer outperforms all other methods with a success rate of 99.67\%, significantly surpassing Dueling DQN (91.17\%) on the 4-section task. As complexity increases in the 10-section task, Powerformer maintains a success rate of 98.19\%, outperforming A2C (80.14\%) by a substantial margin.
For the China 300-bus system,
Powerformer continues to exhibit a satisfying level of performance in comparison to the baseline approaches, accomplishing an impressively high test success rate of $99.73\%$ and $97.36\%$ on the 4-section and 10-section tasks, respectively.  
For the European-9241 system, 
our method consistently demonstrates exceptional performance, achieving an impressive success rate of 68.97\% and 68.25\% when considering the test success rates of 4-section and 10-section tasks.
In contrast, alternative reinforcement learning methods fail to surpass the 35\% threshold for the complex 10-section task. 
Additionally, our method consistently demonstrates its exceptional ability to achieve relatively low economic costs across all scenarios. 



\section{More Detailed Analyses}\label{detailed_analysis}

\subsection{A Discussion from the Perspective of Computer Science}
{
According to the Fig.~\ref{fig:powerformer}, our Powerformer adopts a Transformer based architecture to integrate the section information, which is the the main distinction from the existing architectures. This is the first attempt to use the transformer architecture in the field of transmission section power flow adjustment. 
Specifically, in the existing approaches, the input for state representation learning is homogeneous, while our method can use two heterogeneous electrical information as input. The detailed analyses about section-adaptive attention mechanism, state factorization, GNN integration with transformers, and complexity analysis are as follows:
\\
\noindent
\textbf{Section-adaptive Attention Mechanism.} 
Traditional attention-mask techniques primarily focus on preventing the model from accessing future information during training. However, these techniques typically rely on self-attention, which limit their ability to integrate heterogeneous transmission section information. Our section-adaptive approach differs in that it dynamically optimizes attention distribution based on the current section power flow. This allows for a more sophisticated adaptation to the varying demands of the system, enhancing decision-making for resource allocation.
\\
\noindent
\textbf{State Factorization.} 
The state factorization and the multi-head (MHA) mechanism are different.
The state factors aims to decouple the input into the clear electrical factors (i.e., active power, reactive power, voltage magnitude, and phase angle). They each possess explicit physical meaningful information. This is crucial for understanding the contribution of different factors to the state representation of the power system, which also enhances the interpretability of the achieved representation. As defined in Eq. \ref{mfsa} and Eq. \ref{factor_representation}, our softmax operates between different factors rather than different nodes.
In MHA , the electrical factors are coupling together, making it challenging for each individual attention head to interpret distinct electrical factor significances.
\\
\noindent
\textbf{GNN Integration with Transformers.}
(1) For the task of transmission section power system adjustment, it is often to adjust the generators that are close to the section. Therefore, local information is important for the current section adjustment, and we need to use GNN to capture local information. However, as the power flow is adjusted, the overall operating conditions of the power system changes, and it also affects electrical equipment that is far away. Thus, we need to use Transformer to capture global information. These are essential for developing targeted operational strategies.
(2) By leveraging the capability of GNN, our approach dynamically adapts to topological changes such as maintenance, upgrades, or unforeseen emergencies. For example, when a three-phase short circuit occurs on a transmission line, GNN can accurately capture the changes in local information. This helps to maintain robustness against real-world challenges.\\
\noindent
\textbf{Complexity Analysis.} The time complexity of Powerformer can be analyzed based on Multi-factor Section-adaptive Attention (MFSA) and Graph Neural Network (GNN) propagation. MFSA is ${O}(nd)$, where $n$ represents the number of bus nodes and $d$ represents the dimensionality of the embeddings. On the other hand, GNN propagation is ${O}(nd^2+md)$, where $m$ represents the number of transmission lines. As a result, Powerformer exhibits a time complexity of ${O}(nd+nd^2+md)$, while the traditional Transformer is ${O}(n^2d)$. }

\subsection{A Discussion from the Perspective of Power System}
{
In the field of transmission section power flow adjustment, the use of traditional optimal power flow (OPF) methods cannot consider the complexity of real-world scenarios. Additionally, with the expansion of the power system, OPF also raises concerns regarding real-time performance. Therefore, the deep reinforcement learning (DRL) based  methods has emerged as an innovative alternative for power dispatch. Specifically, our proposed Powerformer demonstrates its effectiveness not only based on the success rate and economic benefits of the dispatch in three cases but also in terms of its fast adjustment time, which meets the demands of real-world scenarios. Moreover, we provide more discussions and analyses about the application in distribution systems, PV systems, embedded distributed storage systems, and the challenge of data drift.
\\
\noindent
\textbf{Distribution Systems.} In distribution systems, our model can be conceptualized as 'feeder-adaptive,' focusing on crucial feeders to control switches that significantly reduce system losses. This adaptation not only serves as load flow adjustment but also demonstrates the model’s applicability in network reconfiguration tasks.
\\
\noindent
\textbf{Photovoltaic (PV) Systems.} For PV systems, our architecture can help in optimizing the flow of generated solar power, which is highly dependent on variable factors such as sunlight availability. The model's adaptive attention mechanisms can dynamically adjust to changes in generation capacity, enhancing the efficiency of power distribution and reducing wastage.
\\
\noindent
\textbf{Embedded Distributed Storage Systems.} In scenarios involving distributed storage systems, our adaptive attention mechanism efficiently identifies and prioritizes the most economically beneficial and supportive energy storage units during complex variations in system operating conditions. By understanding and predicting usage patterns, the system can optimize storage resources to balance the grid during peak and off-peak hours, thereby stabilizing the system and extending battery life.
\\
\noindent
\textbf{Data Drift Challenge.}
In this paper, we focus on the transmission section power flow adjustment, thus not involving data drift. Since data drift is a highly challenging task, which is typically approached as an independent task, involving a detailed analysis and requiring specific designs to effectively address it.
Given the established Powerformer framework of this paper, one solution involves a dataset perspective. This includes creating a training dataset that incorporates a variety of special events to ensure that the data covers a diverse range of scenarios. By doing so, the model can learn from these unique situations, thereby enhancing its robustness and ability to handle real-world complexities effectively.
}

\begin{figure*}
  \centering
  \begin{minipage}{1.0\linewidth}
    \centering
    \subfigure[IEEE 118-bus System]{\includegraphics[width=0.32\textwidth]{figures/ablation.pdf}\label{fig:task10-118}} \hfil
    \subfigure[Realistic 300-bus System]{\includegraphics[width=0.32\textwidth]{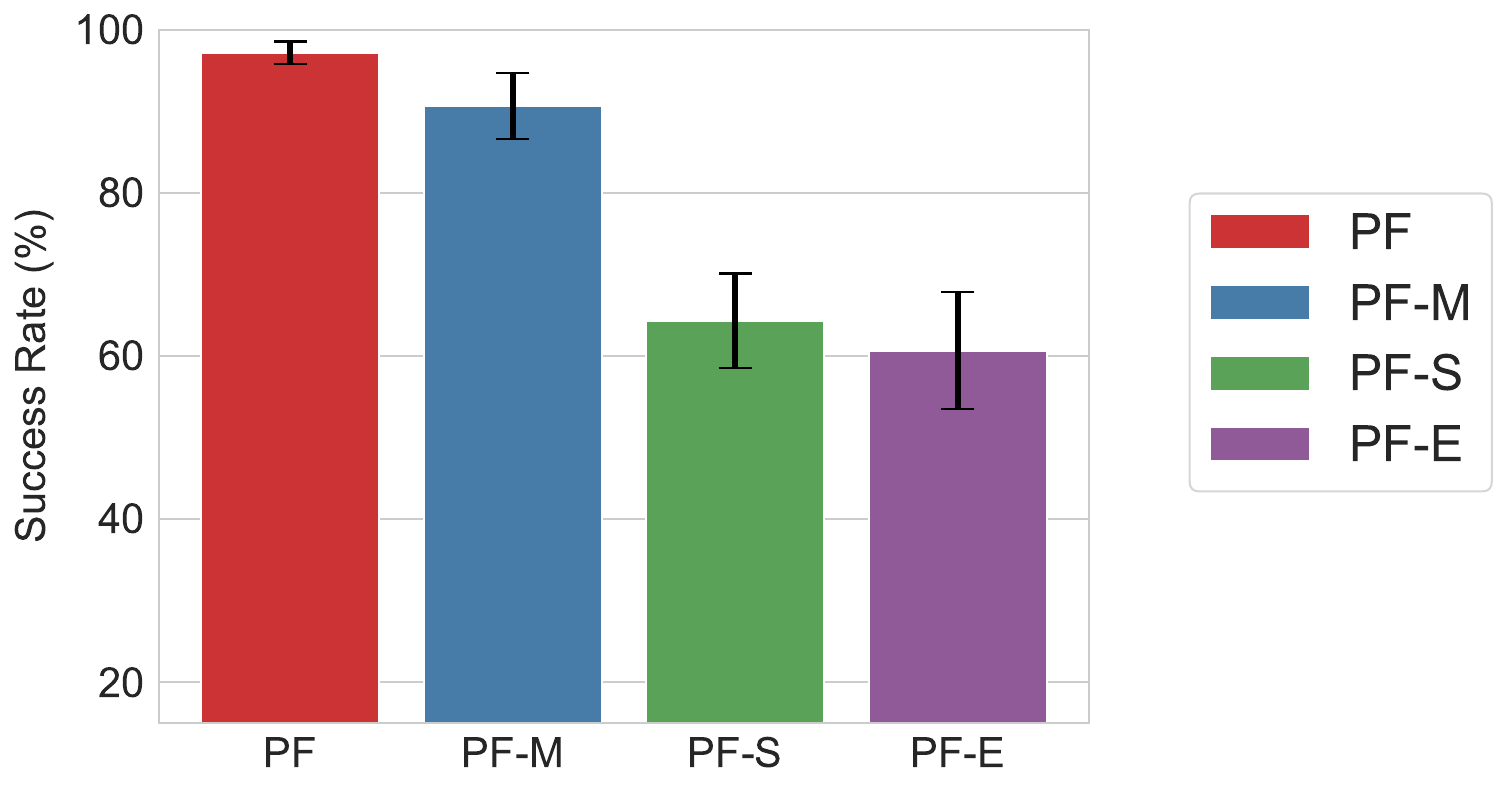}\label{fig:task10-300}} \hfil
    \subfigure[European 9241-bus System]{\includegraphics[width=0.32\textwidth]{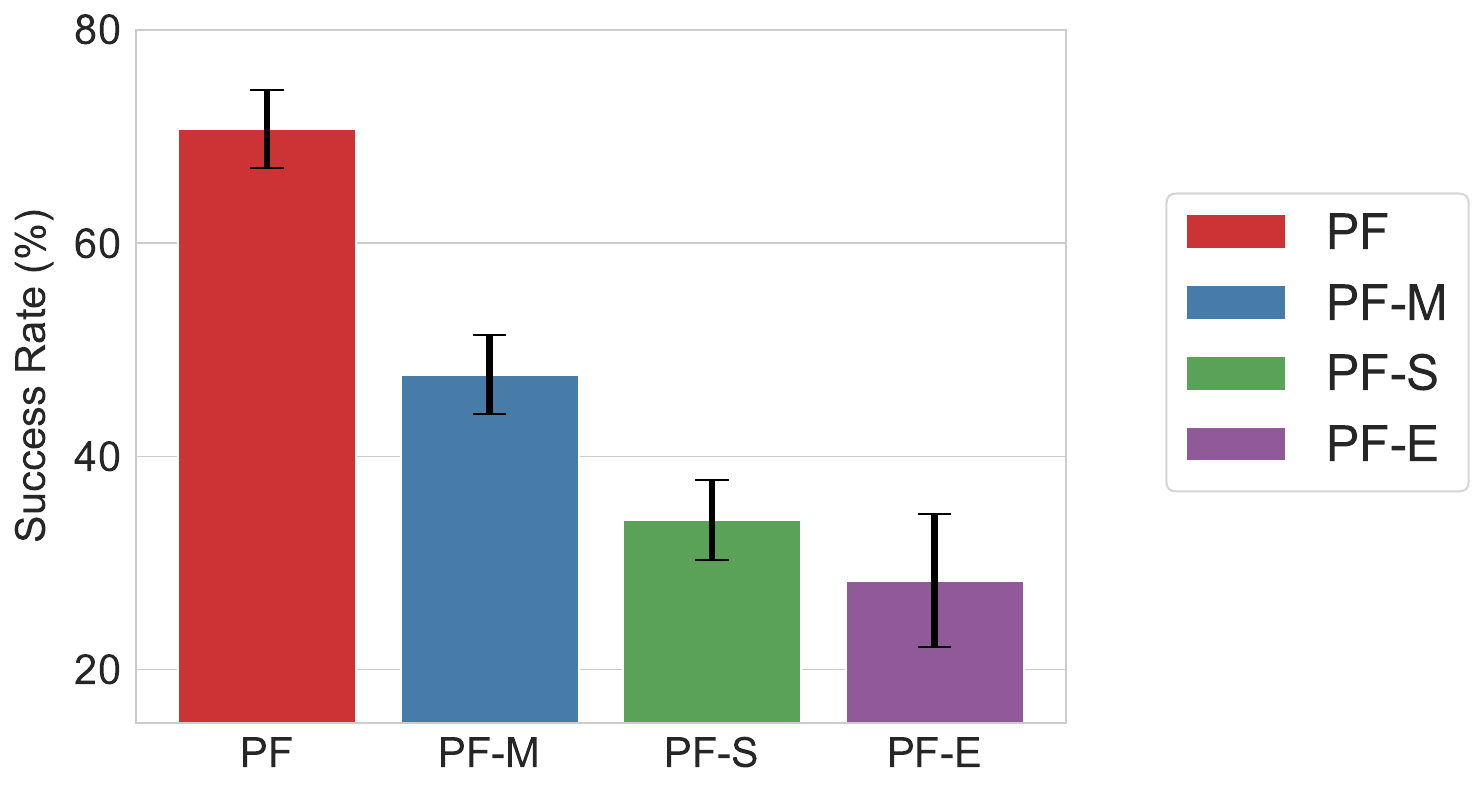}\label{fig:task10-9241}}
    \caption{The performance of our proposed Powerformer and its ablation study on all three power systems.}
    \label{fig:all_ablation}
  \end{minipage}
  \\\vspace{0.2cm}
  
\end{figure*}

\subsection{A comparison with the existing Graph Transformer}

{
The concept of combining GNNs with Transformers has been previously explored in recent years. In this study, we conducted a comparative analysis by considering three classical methods, i.e, Graph Transformer~\cite{yun2019graph}, Graphormer~\cite{ying2021transformers} and GraphFormer~\cite{yang2021graphformers}, to highlight the novelty of our work.
Graph Transformer~\cite{yun2019graph} focuses on node-level prediction within heterogeneous graphs. It utilizes the Transformer's Query, Key, and Value to generate an adjacency matrix. Subsequently, a GNN employs this matrix to learn node representations. Thus, this method does not actually capture global information, as it uses the Q, K, and V solely to learn an adjacency matrix suitable for the task. Graphormer~\cite{ying2021transformers} focuses on graph-level prediction by encoding the structural information of the graph into both the input and attention matrices. It then leverages a Transformer model to learn graph representations. As a result, this method does not use GNN within the framework to capture the local graph structure. Instead, it employs the existing techniques such as centrality encoding in input matrix, alongside spatial and edge encoding in attention matrix. GraphFormer~\cite{yang2021graphformers} treats each node in the graph as an independent sample rather than just a token, and then uses a GNN to aggregate the information processed by the Transformer. Thus, this method does not capture the global information of the entire graph using the transformer, but only focuses on the global level between all tokens within the nodes. Next, we provide the comparison analysis from the perspectives of implementation, effectiveness, and application to specific challenges as follows:
\\
\noindent
\textbf{Implementation.} (1) Graph Transformer employ the Query, Key, and Value to learn adjacency matrix and then use GNN to aggregate the neighborhoods; On the other hand, we employ the Q, K, and V for global information and use GNN to capture the local information. (2) Graphormer embed the graph structure information in the input and attention matrix; On the other hand, we employ GNN as propagation function to capture the local graph information. (3) GraphFormer employ Transformer to capture global level between all tokens within the nodes; On the other hand, we capture global information of the entire graph.
\\
\noindent
\textbf{Effectiveness.} (1) Graph Transformer is demenstrated in node-level prediction within heterogeneous graphs, which is not for this paper's graph-level representation. Thus, it is not inappropriate to obtain a comparative analysis about effectiveness. (2) Graphormer encoding structure information into the input and attention matrices, which requires processing and will incur additional O($n^2$) computational costs; On the hand, our method employ GNN, which is simple and effective, and only adds linear complexity. Moreover, using GNNs is a better choice. If we were to incorporate structural information into the representations like Graphormer does, it would actually introduce noise, as it would affect the meaning of the original electronic factors in the representations. (3) GraphFormer is proposed for node-level prediction within textual graphs, which is not for this paper's graph-level representation. Thus, it is not inappropriate to obtain a comparative analysis about effectiveness.
\\
\noindent
\textbf{Application to specific challenges.} (1) Graph Transformer is designed for heterogeneous, with a specific challenge being the selection of the most important meta-path for GNN. (2) Graphormer faces the challenge of integrating graph structural information into the traditional Transformer architecture. (3) GraphFormer faces the challenge of incorporating textual information within a graph structure. Distinct from these methods, our approach is specifically designed for the power system domain, incorporating transmission section information. The primary challenge here is the effective fusion of section information to maintain accuracy and efficiency. Secondly, we address the complexity of feature entanglement, striving to disentangle into different electronical factors and clarify factor weights for better model performance and insights. Thirdly, to effectively implement dynamic operational strategies in targeted power systems, it is essential to simultaneously utilize both local and global information. This approach allows for precise adjustments to generators close to the section, while also enabling the monitoring and control of electrical equipment located at more distant points within the system.
}

\section{The reason for selecting Dueling DQN}\label{dueling_dqn_selecting}
{
DQN and its variants are off-policy algorithms, while A2C and PPO are on-policy algorithms. Off-policy algorithm enables the agent to update its policy based on the interaction samples from the historical policy, but on-policy algorithm only uses the interaction samples from the current policy. Therefore, the sample efficiency of off-policy algorithm is higher than the on-policy algorithm, which is beneficial in the power system power flow adjustment. In power systems, the environment can be highly stochastic due to unpredictable events such as sudden changes in load or generation. DQN's off-policy nature allows it to effectively learn from older experiences that may capture a variety of these events, whereas PPO may be more focused on recent experiences that might not be as varied.
In the experiments of our original manuscript, the results of 9241-bus System in Table \ref{tab:total} also show that Dueling DQN exhibits superior performances compared to A2C and PPO in the largest power system. In Table \ref{tab:selection_dqn}, we provide experimental comparisons on the IEEE 118 dataset to support our choice of Dueling DQN as the basic backone. }
\begin{table}
  \centering
  \Huge
  \renewcommand{\arraystretch}{1.1} 
  \caption{The comparison of Dueling DQN and A2C with Powerformer on IEEE 118-bus system and China 300-bus systems.}
  \label{tab:selection_dqn}
  \resizebox{0.48\textwidth}{!}{%
  \begin{tabular}{@{}lcccccc@{}}
      \toprule
      \multicolumn{1}{l}{\multirow{2}{*}{{Method}}} 
      & \multicolumn{2}{c}{{118-bus System~(10-section Task)}} 
      & \multicolumn{2}{c}{{300-bus System~(10-section Task)}} 
      
      \\ \cmidrule(l){2-3} \cmidrule(l){4-5} \cmidrule(l){6-7} 
      & \multicolumn{1}{c}{{Success Rate (\%)}} & \multicolumn{1}{c}{{Economic Cost (\$)}} 
      & \multicolumn{1}{c}{{Success Rate (\%)}} & \multicolumn{1}{c}{{Economic Cost (\$)}}  
       \\ \midrule
      A2C + Powerformer              
      & 90.75{\normalsize{~(-7.44)}}     & 626,207{\;\normalsize{~(+4,009)}} & 89.21{\normalsize{~(-8.15)}}
      & 972,523{\;\normalsize{~(+2,841)}}            \\ \specialrule{0em}{1pt}{1pt}
      \textbf{Dueling DQN + Powerformer}              
      & \textbf{98.19}              & {\textbf{622,198}}            & \textbf{97.36} 
      & {\textbf{969,682}}                 \\ \bottomrule
      \end{tabular}
  }
\vspace{-0.5cm}
\end{table}

\section{Action space and reward function}\label{action_space}

\textbf{Action Space} is $\mathcal{A} =\{\sigma_{G_1}^+,\sigma_{G_1}^-,...,\sigma_{G_N}^+,\sigma_{G_N}^-  \}$, where $\mathcal{G}_i$ is the $i$-th controllable generator. Here, $\sigma_G^-$ and $\sigma_G^+$ represent the allowable decrease and increase of generation dispatch. We set $\sigma_G^- = 90\%$ and $\sigma_G^+ = 110\%$ to follow ramp rate limits. If a generator is at its capacity limit, actions causing invalid dispatches are masked to ensure the RL agent stays within operational limits and avoids invalid configurations.

\noindent
\textbf{Reward Function} has two main components:

\noindent
1. Power Flow Component encourages maintaining the power flow within predefined transmission limits. For a single transmission interface $\Phi$, the reward penalizes deviations from the predefined limits as follows: 
\begin{equation}
\mathcal{R}_{pf}(s, a; \Phi) = -|P_{\Phi} - \frac{\mathcal{P}_{\text{min}, \Phi_i} + \mathcal{P}_{\text{max}, \Phi_i}}{2}
  \label{Power_Flow_Component}
\end{equation}

\noindent
2. Economic Dispatch Component is modeled using a quadratic cost function for each generator:
\begin{equation}
\mathcal{R}_{ed}(s, a) = -\sum_{i=1}^{N_G} \left( \alpha_i (P_i^G)^2 + \beta_i P_i^G + \lambda_i \right)
  \label{Economic_Dispatch_Component}
\end{equation}
where $ \alpha_i $, $ \beta_i $, and $ \lambda_i $ are cost coefficients specific to generator  $i$ , and $P_i^G $ is the active power produced by that generator.

\section{More Ablation Studies on Three Datasets}
{To further validate the effectiveness of each component proposed, we conducted ablation experiments on two larger datasets, China 300-bus system and European 9241-bus system. As shown in Fig.~\ref{fig:all_ablation}, the ablation results demonstrate that each component within our Powerformer significantly contributes to the overall performance enhancement. This comprehensive analysis highlights the integral role of each element in achieving the final improvements.
}

\end{document}